\documentclass[sigconf,nonacm,review=false]{acmart}

\usepackage{enumitem}
\usepackage{pifont}
\usepackage{listings}
\usepackage{longtable}
\usepackage{booktabs}
\usepackage{colortbl}
\usepackage{makecell}
\usepackage{siunitx}
\definecolor{winnerviolet}{rgb}{0.94, 0.90, 0.97}
\colorlet{winnercell}{winnerviolet}  % backward-compat alias
\usepackage{tikz}
\usetikzlibrary{arrows.meta,positioning,calc,shapes.geometric,backgrounds,fit}
\usepackage[listings,most]{tcolorbox}

\let\origunderscore\_
\renewcommand{\_}{\origunderscore\linebreak[1]}

\definecolor{codeFrame}{RGB}{90,115,150}
\definecolor{codeBack}{RGB}{245,247,250}
\definecolor{codeString}{RGB}{60,90,130}
\definecolor{codeBrace}{RGB}{70,100,140}

\lstdefinelanguage{json}{
  morestring=[b]",
  stringstyle=\color{codeString},
  morecomment=[l]{//},
  commentstyle=\color{black!45}\itshape,
  literate=
    *{:}{{{\color{black!60}{:}}}}{1}
    {,}{{{\color{black!45}{,}}}}{1}
    {\{}{{{\color{codeBrace}{\{}}}}{1}
    {\}}{{{\color{codeBrace}{\}}}}}{1}
    {[}{{{\color{codeBrace}{[}}}}{1}
    {]}{{{\color{codeBrace}{]}}}}{1},
}

\newtcblisting{jsonbox}[1]{
  listing only,
  listing options={language=json, basicstyle=\ttfamily\scriptsize,
    breaklines=true, breakatwhitespace=false, breakindent=10pt,
    postbreak=\mbox{\hspace{0pt}$\hookrightarrow$\space},
    columns=fullflexible, keepspaces=true, showstringspaces=false},
  colback=codeBack,
  colframe=codeFrame,
  coltitle=white,
  colbacktitle=codeFrame,
  fonttitle=\sffamily\bfseries\small,
  title={#1},
  boxrule=0.5pt,
  arc=2pt,
  left=4pt, right=4pt, top=2pt, bottom=2pt,
  width=\linewidth,
  before=\par\medskip\noindent,
  after=\par\medskip,
  enhanced, breakable,
}

\newtcolorbox{defbox}[2][]{
  colback=white,
  colframe=black!70,
  coltitle=black,
  colbacktitle=white,
  fonttitle=\sffamily\bfseries\small,
  title={#2},
  boxrule=0.6pt,
  arc=0pt,
  left=6pt, right=6pt, top=4pt, bottom=4pt,
  width=\linewidth,
  before=\par\medskip\noindent,
  after=\par\medskip,
  enhanced, breakable,
  attach boxed title to top left={xshift=8pt, yshift=-\tcboxedtitleheight/2},
  boxed title style={colback=white, colframe=white, boxrule=0pt,
                     left=2pt, right=2pt, top=0pt, bottom=0pt},
  #1
}

\newtcblisting{pybox}[1]{
  listing only,
  listing options={language=Python, basicstyle=\ttfamily\scriptsize,
    breaklines=true, breakatwhitespace=false, breakindent=10pt,
    postbreak=\mbox{\hspace{0pt}$\hookrightarrow$\space},
    columns=fullflexible, keepspaces=true, showstringspaces=false,
    keywordstyle=\color{teal!70!black}\bfseries,
    stringstyle=\color{violet!60!black},
    commentstyle=\color{black!50}\itshape,
  },
  colback=teal!4,
  colframe=teal!60!black,
  coltitle=white,
  colbacktitle=teal!60!black,
  fonttitle=\sffamily\bfseries\small,
  title={#1},
  boxrule=0.5pt,
  arc=2pt,
  left=4pt, right=4pt, top=2pt, bottom=2pt,
  width=\linewidth,
  before=\par\medskip\noindent,
  after=\par\medskip,
  enhanced, breakable,
}

\providecommand{\tightlist}{\setlength{\itemsep}{0pt}\setlength{\parskip}{0pt}}

\setcopyright{none}
\acmConference{}{}{}
\acmISBN{}
\acmDOI{}
\acmYear{}
\copyrightyear{}
\renewcommand\footnotetextcopyrightpermission[1]{}
\usepackage{multirow}
\usepackage{tabularray}
\UseTblrLibrary{booktabs}
\graphicspath{{figures/}}

\definecolor{darkgreen}{RGB}{0, 181, 18}
\definecolor{DeltaBg}{HTML}{D4F2D7}
\definecolor{SearchBg}{HTML}{C2E6F5}
\definecolor{AgenticBg}{HTML}{F5C2CC}
\definecolor{MathBg}{HTML}{E6D4F2}
\definecolor{ScienceBg}{HTML}{FBE0BC}
\definecolor{tblHeaderTint}{RGB}{220,228,240}
\definecolor{tblRowAlt}{RGB}{246,246,248}
\newcommand{\gain}[1]{$\uparrow$ #1}

\title{Invalidation Contracts for Cross-Episode Agent Memory}

\author{Michael Wu}
\authornote{Both authors contributed equally to this work.}
\affiliation{%
  \institution{South Dakota State University}
  \city{Brookings}
  \state{SD}
  \country{USA}}
\email{Wu.Beining@jacks.sdstate.edu}

\author{Arquimedes Canedo}
\authornotemark[1]
\affiliation{%
  \institution{Siemens Digital Industries Software}
  \city{Princeton}
  \state{NJ}
  \country{USA}}
\email{arquimedes.canedo@siemens.com}

\begin{document}

\begin{abstract}
LLM agents that cache recovery suggestions from API errors can skip
re-derivation in later episodes, spending fewer tokens and fewer model
calls on constraints they have already learned. Server-side data drift
turns those cached fixes into silent failures, and the usual remedy,
re-deriving on every episode, gives the savings back. We introduce
\emph{invalidation contracts}, a protocol layer that attaches version
stamps and cacheability hints to every recovery suggestion so the
client can evict stale entries without trial and error, and keep the
rest. The contract decomposes realized savings into two independent
factors: \emph{validity}, the fraction of cached suggestions that
remain correct after a drift event, and \emph{compliance}, the
fraction the planner applies on the first attempt. Validity depends
only on the protocol and is vendor-independent. Compliance depends on
the planner model: identical wire bytes yield 100\% first-try
compliance on Claude Haiku~4.5 and 11\% or below on Claude Sonnet~5, which
exhibits \emph{input-schema conservatism}, refusing fixes that add
fields the original request did not contain. We evaluate across seven
models, three serving paths, two domains, and approximately 9,400
episodes. Row-level invalidation raises compliance by 0 to 66.7
percentage points across the seven models, 55.6 to 66.7 on three, and
recovers 29--33\% of baseline token cost on four of seven models,
while table-level invalidation destroys co-located entries and drops
post-drift first-try rates to 0\% on five of seven. Eviction precision
is 1.00 at row granularity on every model under the row-level oracle
of \S\ref{sec:hygiene}. The contract adds 15\% to response payload.
Version-stamp validity is deterministic by construction and produced
identical results across every model and serving path, with zero
contract failures in the entire evaluation.

\end{abstract}


\maketitle

\section{Introduction}
\label{sec:introduction}

LLM agents that call APIs repeatedly tend to remember what worked and
reuse it in later episodes~\cite{shinn2023reflexion,wang2023voyager}.
This saves tokens when the server's reference data stays the same.
When the data changes, the cached fix becomes wrong. Naive memory
(caching without invalidation) saves 5--28\% of tokens over no memory
at all, but table-level invalidation destroys that gain on every
constraint class that shares a table with the one that
drifted: post-drift first-try rates drop to 0\% on five of seven
models.\footnote{All opening numbers trace to the A1 and A2 arms of
  the drifted stream in Table~\ref{tab:ladder} and
  Table~\ref{tab:tokens}.}
Imprecise invalidation is actively harmful.

The problem has a structural root. Recovery suggestions today~\cite{canedo2026selfreflective} carry no
metadata about their validity. An agent that caches a suggestion has
no signal to tell it when the suggestion expires. It can only discover
staleness by replaying the error and watching the fix fail. HTTP data
responses solved this decades ago with \texttt{Cache-Control},
\texttt{ETag}, and conditional requests~\cite{rfc7234}. The IETF
tradition standardized the diagnostic envelope for API
errors~\cite{rfc7807,rfc9457} but left the recovery layer empty. No
protocol tells the agent what to cache, for how long, or when to
revalidate.

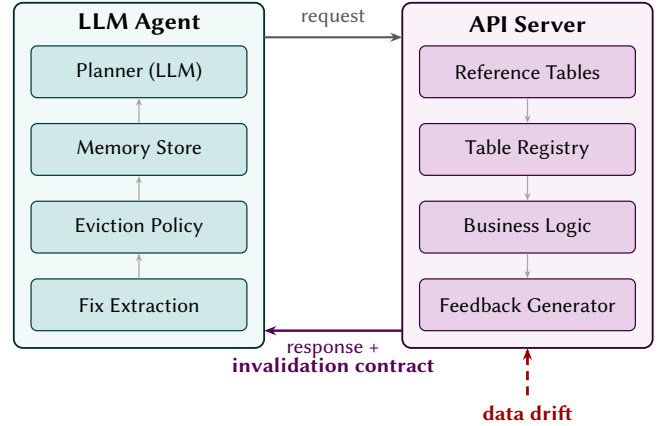
\begin{figure}[t]
\centering
\resizebox{\columnwidth}{!}{%
\begin{tikzpicture}[
  arr/.style={-{Stealth[length=4pt,width=3pt]}, line width=0.7pt},
  iarr/.style={-{Stealth[length=2.5pt,width=2pt]}, line width=0.4pt,
               black!35},
  sbox/.style={rounded corners=2pt, fill=violet!18,
               draw=violet!45!black, line width=0.4pt,
               font=\sffamily\footnotesize,
               minimum height=0.6cm, minimum width=2.5cm,
               inner sep=2pt},
  cbox/.style={rounded corners=2pt, fill=teal!18,
               draw=teal!45!black, line width=0.4pt,
               font=\sffamily\footnotesize,
               minimum height=0.6cm, minimum width=2.5cm,
               inner sep=2pt},
  heading/.style={font=\sffamily\small\bfseries},
]
% --- Client frame ---
\draw[rounded corners=3pt, fill=teal!5, draw=teal!35!black,
      line width=0.6pt]
  (-1.45,-1.85) rectangle (1.45,2.15);
\node[heading] at (0,1.9) {LLM Agent};

\node[cbox] (plan) at (0, 1.35) {Planner (LLM)};
\node[cbox] (mem)  at (0, 0.45) {Memory Store};
\node[cbox] (evic) at (0,-0.45) {Eviction Policy};
\node[cbox] (fix)  at (0,-1.35) {Fix Extraction};

\draw[iarr] (fix)  -- (evic);
\draw[iarr] (evic) -- (mem);
\draw[iarr] (mem)  -- (plan);

% --- Server frame ---
\draw[rounded corners=3pt, fill=violet!5, draw=violet!35!black,
      line width=0.6pt]
  (3.05,-1.85) rectangle (5.95,2.15);
\node[heading] at (4.5,1.9) {API Server};

\node[sbox] (ref)  at (4.5, 1.35) {Reference Tables};
\node[sbox] (reg)  at (4.5, 0.45) {Table Registry};
\node[sbox] (biz)  at (4.5,-0.45) {Business Logic};
\node[sbox] (fb)   at (4.5,-1.35) {Feedback Generator};

\draw[iarr] (ref) -- (reg);
\draw[iarr] (reg) -- (biz);
\draw[iarr] (biz) -- (fb);

% --- Request arrow ---
\draw[arr, black!65] (1.45,1.75)
  -- node[above, font=\sffamily\footnotesize] {request} (3.05,1.75);

% --- Response arrow ---
\draw[arr, violet!65!black, line width=0.8pt]
  (3.05,-1.65)
  -- node[below, font=\sffamily\footnotesize, align=center,
          yshift=-1pt]
     {response +\\[-2pt]\textbf{invalidation contract}}
  (1.45,-1.65);

% --- Data drift ---
\draw[arr, red!60!black, line width=0.9pt, densely dashed]
  (4.5,-2.4) -- (4.5,-1.85);
\node[font=\sffamily\footnotesize\bfseries, red!60!black]
  at (4.5,-2.6) {data drift};
\end{tikzpicture}%
}
\caption{System overview.
  On the client (left), the planner uses cached fixes to generate the
  next request; the response comes back to fix extraction, the
  eviction policy governs the memory store, and the store feeds the
  planner again. On the server (right), reference tables feed a
  versioned registry, which feeds business-logic policies, which feed
  the feedback generator. The feedback generator attaches
  invalidation metadata to every response. Data drift enters the
  server through the reload endpoint.}
\label{fig:system}
\end{figure}

Figure~\ref{fig:system} shows the architecture. An LLM agent calls an API
server repeatedly across episodes. Between episodes the agent may
cache recovery suggestions from previous failures and inject them as a
memory block into the next request. The server may update its reference
data at any time (data drift). The invalidation contract is the set of fields the
server attaches to every response so the client can decide what to keep
and what to evict.

This paper introduces \textbf{invalidation contracts}, a protocol
layer that attaches cache-control semantics to recovery suggestions.
Two fields on every suggestion (a cacheability hint and a version
stamp) plus a structured diff on schema reload give the agent enough
information to evict stale knowledge without trial and error. We
implement six protocol levels, from a bare version stamp through
row-level diffs to dependency-vector comparison, and measure each
across seven models, three serving paths, two domains, and
approximately 9{,}400 episodes.

The contract splits realized savings into two factors:

\begin{defbox}{The Savings Equation}
\vspace{-0.5em}
\begin{equation}
\label{eq:savings}
\text{realized savings} = \text{validity} \times \text{compliance}(m,\,p,\,a)
\end{equation}
\vspace{0.3em}
\noindent\small $m$\,=\,model,\quad $p$\,=\,protocol,\quad $a$\,=\,action type
\end{defbox}
\emph{Validity} is the fraction of cached suggestions that remain
correct after a drift event. The contract controls it entirely.
Validity is identical across all seven models and all three serving
paths we tested, with zero contract failures in approximately 9{,}400
episodes.
\emph{Compliance} is the fraction of valid suggestions the agent
actually applies on the first attempt. It depends on the LLM
model~($m$), the protocol level~($p$, our Levels 0--6), and the
action type~($a$, whether the fix rewrites an existing field or adds
a new one). Compliance varies by model but is engineerable. Adding
row-level invalidation raised compliance by 0 to 66.7 percentage
points across the seven models, and by 55.6 to 66.7 on three of them,
without changing the model.\footnote{A2D over A1 in
  Table~\ref{tab:ladder}: Claude Haiku~4.5 (+55.6\,pp),
  Claude Sonnet~4.6 (+63.0\,pp), GPT-5-mini (+66.7\,pp);
  \texttt{gemini-3.5-flash} (+48.2\,pp), Claude Sonnet~5 (+11.1\,pp),
  and \texttt{gpt-5.4-mini} and \texttt{deepseek-v4-flash} (0.0\,pp),
  the last two already at their compliance ceiling and floor
  respectively at A1.}

We make the following contributions:
\begin{enumerate}[leftmargin=*,nosep]
  \item The \textbf{invalidation contract} protocol, with six measured
        levels from version stamps through dependency vectors
        (\S\ref{sec:contract}).
  \item Empirical evidence across seven models, three serving paths,
        two domains, and ${\sim}$9{,}400 episodes with zero contract
        failures, showing that validity is a protocol property and
        compliance is a model property (\S\ref{sec:results}).
  \item A savings decomposition that separates the term the API
        provider controls (validity) from the term the model vendor
        controls (compliance), giving API providers a concrete
        engineering target
        (\S\ref{sec:results}).
  \item A price sheet for eager versus lazy rule-drift detection, with
        both columns measured on the same
        stream (Table~\ref{tab:pricesheet}, \S\ref{sec:disc-detection}).
  \item \textbf{Input-schema conservatism}: the observation that a
        model may refuse to add a field its input schema did not
        contain even when told exactly which field to add, which
        bounds what any protocol can achieve and is not predicted by
        model recency (\S\ref{sec:disc-reach}).
\end{enumerate}

\noindent
The server implementation is available at
\url{https://github.com/arquicanedo/self-reflective-apis} and the
client-side evaluation harness at
\url{https://github.com/beining1008/cross-episode-memory-client}.
All reported numbers trace to the archived data layer in the client
repository: the per-run and per-cell tab-separated (TSV) extraction
behind Tables~\ref{tab:ladder}--\ref{tab:taxonomy}, and the
per-episode run artifacts with vendor token-accounting fields
removed.

\section{Related Work}
\label{sec:related_work}

\paragraph{Storing what worked.}
\looseness=-1
Deployed agents write successful calls and repairs into a persistent store and reuse them in later
episodes~\cite{shinn2023reflexion, wang2023voyager, liang2023taskmatrix, mialon2023augmented}. The
interface those calls travel over has been standardized in parallel, from tool-invocation
frameworks~\cite{qin2023toolllm} to the vendor and agent protocols now carrying
them~\cite{openai2025functioncalling, mcp2025}, and benchmarks evaluate the resulting agents against
API suites, web environments and repository tasks~\cite{li2023apibank, chen2023teval,
liu2023agentbench, zhou2024webarena, yang2023intercode, yang2024sweagent}. What this line governs is
what enters the store and how it is retrieved. An entry's validity is settled once, at the moment it
is written, and nothing in the interface states when it stops holding.

\paragraph{Verdicts the client computes for itself.}
\looseness=-1
A natural remedy is to have the client audit the stored entry, and the record separates two
settings. Absent an external signal, self-correction does not reliably improve an answer, and
self-critique scores track the model's confidence rather than the world~\cite{huang2023cannotcorrect,
kamoi2024selfcorrection, stechly2023gpt4wrong, valmeekam2023selfcritique}, though iterative
refinement remains useful where the text is itself the objective~\cite{madaan2023selfrefine,
paul2023refiner}. The second setting supplies an external signal: a compiler, a test, or a tool
response. There repair works, and the gain belongs to the signal rather than to
introspection~\cite{gou2024critic, chen2023selfdebugging, olausson2024selfrepair}. Recent
benchmarks now measure that separation directly~\cite{le2026pairbench, dai2026feedbackeval,
sriram2026multitool}. Both settings compute the
verdict from what the client already holds. A reference table that rotated on the server leaves no
trace in that context, so no amount of self-checking recovers it.

\paragraph{Pricing freshness on the wire.}
\looseness=-1
Deciding what a caller may keep, and when it must check again, is old work. HTTP gives a response an
explicit freshness lifetime and a validator the caller presents to revalidate without
refetching~\cite{rfc7234}. Service description carries the same discipline into APIs, from semantic
markup~\cite{owls2004, wsmo2005} through hypermedia constraints~\cite{fielding2008hateoas,
fowler2010maturity} to machine-readable schemas and typed introspection~\cite{openapi2021,
graphql2021}, and the agent-facing protocols inherit it~\cite{yang2025agentprotocols}. The error
path was standardized separately, as a typed envelope reporting what went wrong~\cite{rfc7807,
rfc9457}, and structured recovery suggestions sit one layer above
it~\cite{canedo2026selfreflective}. In all of it the cache-control vocabulary attaches to the data
response and the diagnostic vocabulary attaches to the error, while the suggestion, the one object
an agent actually caches, carries neither. The invalidation contract puts both on that object: a
hint saying whether the suggestion may be kept at all, and a stamp saying when it stopped being
true.

\section{The Invalidation Contract}
\label{sec:contract}

An invalidation contract is a set of machine-readable fields that travel
with every API response, telling the caller what it may cache, what that
cache depends on, and when the dependency has changed.

% ------------------------------------------------------------------
\subsection{The Two APIs}
\label{sec:domains}

The contract fields in the rest of this section name concepts that
belong to the APIs underneath, not to the protocol.  We define them
first, because the listings are unreadable otherwise.

\begin{figure}[t]
\centering
\resizebox{\columnwidth}{!}{%
\begin{tikzpicture}[
  arr/.style={-{Stealth[length=4pt,width=3pt]}, line width=0.7pt},
  darr/.style={-{Stealth[length=3pt,width=2.5pt]}, line width=0.5pt,
               black!45, densely dashed},
  tbox/.style={rounded corners=2pt, fill=violet!18,
               draw=violet!45!black, line width=0.4pt,
               font=\sffamily\footnotesize,
               minimum height=0.6cm, minimum width=2.6cm,
               inner sep=2pt},
  % Everything in this figure is server-side, so both box styles stay in
  % the violet family; teal is the client colour in Figure~\ref{fig:system}.
  pbox/.style={rounded corners=2pt, fill=violet!7,
               draw=violet!45!black, line width=0.4pt,
               font=\sffamily\footnotesize,
               minimum height=0.6cm, minimum width=2.6cm,
               inner sep=2pt},
  lbl/.style={font=\sffamily\scriptsize, black!65},
  heading/.style={font=\sffamily\small\bfseries},
]
% --- generic model, left ---
\node[heading] at (0,2.5) {Domain model};

\node[tbox] (tbl)  at (0, 1.85) {Reference table \textsf{v}};
\node[tbox] (row)  at (0, 1.00) {Row (key $\rightarrow$ values)};
\node[pbox] (rule) at (0, 0.15) {Derivation rule};
\node[pbox] (pol)  at (0,-0.70) {Policy};
\node[pbox] (sug)  at (0,-1.55) {Recovery suggestion};

\draw[arr] (tbl)  -- node[right, lbl] {contains} (row);
\draw[arr] (row)  -- node[right, lbl] {related by} (rule);
\draw[arr] (rule) -- node[right, lbl] {enforced by} (pol);
\draw[arr] (pol)  -- node[right, lbl] {violated $\Rightarrow$} (sug);

% --- Acme instantiation, right ---
\node[heading] at (4.9,2.5) {Acme billing instance};

\node[tbox] (t1) at (4.9, 1.85)
  {\texttt{active\_csm\_codes} 1.0.0};
\node[tbox] (t2) at (4.9, 1.00)
  {\texttt{plan\_partner\_growth}};
\node[pbox] (t3) at (4.9, 0.15)
  {\texttt{requires\_recognition}};
\node[pbox] (t4) at (4.9,-0.70)
  {promo must be eligible};
\node[pbox] (t5) at (4.9,-1.55)
  {\texttt{USE\_REQUIRED\_PROMO}};

\draw[arr] (t1) -- (t2);
\draw[arr] (t2) -- (t3);
\draw[arr] (t3) -- (t4);
\draw[arr] (t4) -- (t5);

\draw[darr] (tbl)  -- (t1);
\draw[darr] (row)  -- (t2);
\draw[darr] (rule) -- (t3);
\draw[darr] (pol)  -- (t4);
\draw[darr] (sug)  -- (t5);

% --- drift ---
\draw[arr, red!60!black, line width=0.8pt, densely dashed]
  (7.0,1.85) -- (6.25,1.85);
\node[font=\sffamily\scriptsize\bfseries, red!60!black, align=center]
  at (7.9,1.85) {drift bumps\\the version};
\end{tikzpicture}%
}
\caption{The domain model both APIs share, with the Acme billing
  instantiation beside it. A versioned reference table holds keyed
  rows; derivation rules relate rows across tables; policies enforce
  the rules; a violated policy emits a recovery suggestion. Data
  drift changes a table and bumps its version, which is the event the
  invalidation contract exists to report. The Level~5 listing's graph
  nodes are the \texttt{table:key} pairs of the second row.}
\label{fig:domain-model}
\end{figure}
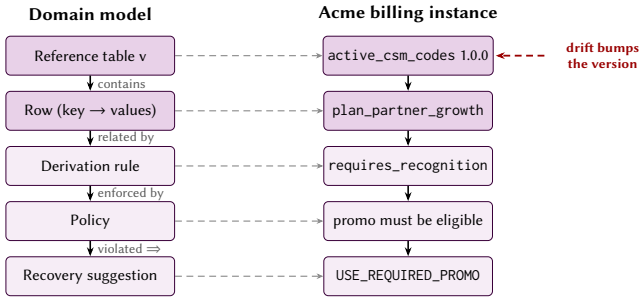

Both APIs share one domain model, shown in
Figure~\ref{fig:domain-model}.  A \emph{reference table} is a named,
server-side dictionary from a \emph{key} to a \emph{row} of
values: \texttt{token\_funding\_map}, for instance, maps a payment
token to its funding type.  Each table carries a \emph{version}, a
string the server bumps whenever the table's contents change.  A
\emph{derivation rule} relates one table to another: a declaration
that a value in table $A$ must correspond to a value in table $B$.
The set of rule declarations has its own fingerprint, which changes
when rules are added or redeclared but not when table data changes.  A
\emph{policy} is a validation the API enforces on an incoming
request by consulting one or more tables through zero or more rules.
When a request violates a policy, the server rejects it and emits a
\emph{recovery suggestion}: a typed, machine-readable description of
the change that would make the request pass.  Everything the contract
adds is metadata about which tables and rules a given suggestion
depended on.

\paragraph{Acme billing.}
The primary domain wraps Stripe, a commercial payment-processing
service, with five policies over six reference tables.  Three tables
carry the paper's examples.  \texttt{active\_csm\_codes} maps a plan
to the promotional code currently assigned to it by a customer success
manager (CSM), the human account owner; these codes rotate on a
business schedule, which is the paper's breaking drift event.
\texttt{promo\_eligibility} records which promotional codes a plan may
actually redeem, and the rule \texttt{requires\_recognition} ties the
two together: a code assigned in the first table must exist in the
second.  \texttt{token\_funding\_map} maps a payment-method token to
its funding type (credit or debit), and
\texttt{recommended\_credit\_token} names the replacement to use when
a policy requires credit.  These two are related by the funding-type
policy, which is why a single funding fix depends on two tables, the
case Level~4 exists to express.

\paragraph{Recipe conversion.}
The second domain validates ingredient substitutions against reference
tables of celiac-safe brands and incompatible ingredient
combinations.  Its structure is the same, one rung simpler: policies
consult tables, but the dietary ladder's rules chain less deeply.  It
is in the evaluation to supply a third hint shape, a
\emph{rewrite-a-value} hint that replaces an ingredient the task
already names, and to test whether the contract's behaviour survives a
change of domain.  Both APIs share the same invalidation contract.
The drift schedule and the episode streams are in \S\ref{sec:setup}.

% ------------------------------------------------------------------
\subsection{Protocol Elements}
\label{sec:protocol-elements}

The contract adds three per-suggestion fields and two response-level
fields.  All are optional: a server that emits none of them produces
a Level~0 response (plain-text error, nothing cacheable).

\paragraph{Per-suggestion fields.}
Every recovery suggestion carries:

\begin{itemize}\tightlist
\item \texttt{cache\_hint}: \texttt{"cacheable"} or
  \texttt{"recompute"}.  A cacheable fix comes from a closed,
  server-side reference table (e.g., a mapping from payment-method
  tokens to funding types); a recompute fix is derivable from the
  request itself (e.g., a numeric rounding correction).
\item \texttt{table}: the name of the reference table the fix depends
  on.  Present only when \texttt{cache\_hint} is \texttt{"cacheable"}.
\item \texttt{tables}: a dependency vector, a list of
  \texttt{\{table, version\}} pairs for fixes that depend on more than
  one table.  When present, \texttt{tables} takes precedence over the
  scalar \texttt{table} field.  This is a Level~4 addition
  (\S\ref{sec:protocol-levels}).
\end{itemize}

\paragraph{Response-level fields.}
Every response (success or error) carries:

\begin{itemize}\tightlist
\item \texttt{table\_versions}: a dictionary mapping each loaded
  reference table to its current version string.  The client compares
  this against the versions it stored at cache time; a mismatch on any
  table referenced by a cached suggestion triggers invalidation.
\item \texttt{rules\_version}: a 12-character SHA-256 prefix of the
  current derivation-rule declarations (from-table, to-table, relation
  triples).  This hash changes when rules are added, removed, or
  redeclared with a different relation, but not when table data changes.
  It enables eager detection of rule drift on the \emph{success} path,
  where no recovery suggestions are emitted and per-suggestion fields
  are absent.
\end{itemize}

The contract has two channels, and the ladder in
\S\ref{sec:protocol-levels} interleaves them because that is the order
in which we found them to matter. The \emph{wire} channel is what the
server emits: Levels~0, 1, 4, 5, and~6 each add fields to the previous
level's response. The \emph{consumption} channel is how the client
presents a fix to the planner once it has arrived: Levels~2 and~3
change the system prompt and the memory block, and leave the wire
format byte-for-byte identical to Level~1.

That distinction matters for reading the rest of this section. A
consumption level cannot have a listing, because there is nothing new
on the wire to show. It is still a level in the sense that it is an
independent, orderable intervention with a measured effect on realized
savings, but it is not a change an API provider ships. Both are shown
in Table~\ref{tab:protocol-levels}, which marks each level with where
it takes effect. \S\ref{sec:consumption-levels} describes what
Levels~2 and~3 put in the prompt and why. The listings that follow
show the wire additions at Levels~1, 4, 5, and~6.

\begin{jsonbox}{Level 1: structured suggestion with version stamps}
{
  "success": false,
  "recovery_feedback": {
    "suggestions": [
      {
        "type": "USE_REQUIRED_FUNDING_TYPE",
        "parameters": {"payment_method_token": "acme_pm_visa_credit"},
        "cache_hint": "cacheable",
        "table": "recommended_credit_token"
      }
    ]
  },
  "table_versions": {
    "recommended_credit_token": "1.0.0",
    "token_funding_map": "1.0.0",
    "active_csm_codes": "1.0.0"
  }
}
\end{jsonbox}

\noindent
Level~1 is the root of the contract. The client caches the suggestion
and tags it with \texttt{table}~=~\texttt{"recommended\_credit\_token"}
and the corresponding version \texttt{"1.0.0"} from
\texttt{table\_versions}. On a later response, if
\texttt{table\_versions} reports a different version for that table,
the client evicts the entry.

\begin{jsonbox}{Level 4: adds dependency vector (suggestion excerpt)}
{
  "type": "USE_REQUIRED_FUNDING_TYPE",
  "parameters": {"payment_method_token": "acme_pm_visa_credit"},
  "cache_hint": "cacheable",
  "table": "recommended_credit_token",
  // Level 4 addition: both tables this fix depends on
  "tables": [
    {"table": "token_funding_map", "version": "1.0.0"},
    {"table": "recommended_credit_token", "version": "1.0.0"}
  ]
}
\end{jsonbox}

\noindent
The funding-type fix depends on two tables:
\texttt{token\_funding\_map} determines whether the payment method's
funding type is permitted, and \texttt{recommended\_credit\_token}
provides the replacement token. Level~4 makes both dependencies
explicit in the \texttt{tables} list. The client invalidates when
either version moves, without needing to infer the relationship
itself.

\begin{jsonbox}{Level 5: knowledge graph (GET /admin/graph, excerpt)}
// 23 nodes and 20 edges total; two shown here
{
  "nodes": {
    "active_csm_codes:plan_partner_growth": {
      "table": "active_csm_codes",
      "key": "plan_partner_growth", "version": "1.0.0"
    },
    "promo_eligibility:SUMMERSALE25": {
      "table": "promo_eligibility",
      "key": "SUMMERSALE25", "version": "1.0.0"
    }
  },
  "edges": [
    { "from": "active_csm_codes:plan_partner_growth",
      "to":   "promo_eligibility:SUMMERSALE25",
      "relation": "requires_recognition" }
  ]
}
\end{jsonbox}

\noindent
The edge \texttt{requires\_recognition} encodes that the CSM code
assigned to \texttt{plan\_partner\_growth} must exist in
\texttt{promo\_eligibility}. When the CSM table rotates from
\texttt{SUMMERSALE25} to \texttt{WINTERLAUNCH26}, the server walks
this edge and reports
\texttt{promo\_eligibility:SUMMERSALE25} as
\texttt{affected\_downstream}. The client can query the full graph
(23~nodes, 20~edges in the Acme domain) via
\texttt{GET~\!/admin/graph}. The walk uses the pre-reload graph
snapshot because the edge to \texttt{SUMMERSALE25} disappears after
the new data is applied.

\begin{jsonbox}{Level 6: rules fingerprint on a success response}
// No recovery_feedback on success, but the
// fingerprint still travels with the response
{
  "success": true,
  "table_versions": {
    "recommended_credit_token": "1.0.0",
    "token_funding_map": "1.0.0",
    "active_csm_codes": "1.0.0"
  },
  "rules_version": "ec524f499ebe"
}
\end{jsonbox}

\noindent
This is a success response with no \texttt{recovery\_feedback}. The
client never sees dependency vectors on the success path because they
ride only on suggestions. The \texttt{rules\_version} hash
(\texttt{ec524f499ebe}) fills that gap. If a later response carries a
different hash while all \texttt{table\_versions} stay the same, the
client knows a derivation rule changed and can compare its cached
vectors against the new schema on the next failure.

\paragraph{Reload endpoint.}
A \texttt{POST /admin/reload-table} endpoint in the
server lets an
operator push new table data without restarting the server. In the
client-side evaluation harness, the \texttt{DriftDriver} automates this by reading declarative
scenario manifests that specify which tables change at which episode
and from which snapshot.  The response includes the
old and new version strings, a structured diff (added, removed, and
changed keys), and a list of \texttt{affected\_downstream} node
identifiers computed by walking the knowledge graph forward from the
changed keys.  The walk uses the \emph{pre-reload} graph snapshot:
edges from the old data (which the client still holds in its cache) may
disappear after the reload, so the server must traverse them before
overwriting.

% ------------------------------------------------------------------
\subsection{Protocol Levels}
\label{sec:protocol-levels}

Table~\ref{tab:protocol-levels} arranges the contract into seven
levels along the two axes just introduced.  The ``Where'' column
records which channel each level acts on: five levels change the wire
format, and two change only how the client consumes what arrives.
L0--L3 address \emph{compliance}: getting the planner to apply the fix
at all.  L4--L6 address \emph{validity}: ensuring a cached fix still
matches the current server state.  L1 straddles both factors: its
structured suggestion format aids compliance, while its version stamps
and \texttt{cache\_hint} enable invalidation.  The experiments in
\S\ref{sec:results} hold compliance fixed at L3 (which subsumes L2)
and vary the validity channel across the A-ladder
(\S\ref{sec:two-ladders}).

\begin{table*}[t]
\centering
\caption{Levels of the invalidation contract.  ``Where'' marks the
  channel: \emph{wire} levels change what the server emits;
  \emph{consumption} levels change how the client presents an arrived
  fix to the planner, leaving the wire format identical to L1.  Only
  wire levels are something an API provider ships.  L0--L3 engineer
  compliance; L4--L6 engineer validity.  L1 straddles both: its
  structured format aids compliance, and its version stamps enable
  invalidation.  This evaluation holds compliance fixed at L3 (which
  subsumes L2) and varies the validity channel; see
  Table~\ref{tab:arm-levels} for the arm-to-level mapping.}
\label{tab:protocol-levels}
\small
\begin{tblr}{
  colspec = {c l c X[1.0] X[1.1]},
  width = \textwidth,
  rowsep = 3pt,
  row{1} = {font=\bfseries, bg=tblHeaderTint},
  row{3,5,7} = {bg=tblRowAlt},
  hline{1,Z} = {0.8pt},
  hline{2} = {0.5pt},
}
Level & Name & Where & What changes & What it solves \\
L0 & Raw feedback~\cite{canedo2026selfreflective}
  & Wire
  & Plain-text error
  & Nothing cacheable \\
L1 & Structured suggestion
  & Wire
  & Typed fix + \texttt{cache\_hint} + \texttt{table} + version
  & Single-table invalidation \\
L2 & Patch protocol
  & Consumption
  & Wire as L1; system prompt teaches ``merge parameters verbatim''
  & Planner compliance \\
L3 & Amended input
  & Consumption
  & Wire as L1; memory block presents fixes as pre-applied amendments
  & Removes instruction conflict (minor nudge) \\
L4 & Dependency vectors
  & Wire
  & \texttt{tables}: [\{table, version\}, \ldots] per suggestion
  & Multi-table invalidation without client-side derivation \\
L5 & Subgraph propagation
  & Wire
  & Versioned graph: nodes = facts, edges = derivations
  & Chain invalidation at arbitrary depth \\
L6 & Rules fingerprint
  & Wire
  & \texttt{rules\_version} hash on every response
  & Eager detection of rule drift on the success path \\
\end{tblr}
\end{table*}

The wire-format delta between levels is small.  At Level~1 each
suggestion carries two fields (\texttt{cache\_hint} and \texttt{table})
and the response envelope carries \texttt{table\_versions}.  Level~4
adds one field per suggestion (\texttt{tables}, a list of
\texttt{\{table, version\}} pairs).  Level~6 adds one field to the
envelope (\texttt{rules\_version}).  The full Level~4 response is shown
in \S\ref{sec:protocol-elements}; the incremental fields at each level
are visible in the column ``What changes'' of
Table~\ref{tab:protocol-levels}.

% ------------------------------------------------------------------
\subsection{The Consumption Levels}
\label{sec:consumption-levels}

Levels~2 and~3 are the two consumption levels.  Neither adds a byte to
the wire; both change what the client puts in front of the planner
after a Level~1 response arrives.  We state what each one writes,
because a level with no listing is otherwise unreproducible.

\paragraph{Level 2: patch protocol.}
The client adds one sentence to the system prompt: ``merge the
suggestion parameters into your next request verbatim; do not
paraphrase or omit fields.''  Without it, a planner that has read a
structured suggestion still tends to restate the fix in its own
vocabulary, renaming a field or dropping one it judges redundant, and
the server rejects the retry for a second reason.  The sentence
converts the suggestion from advice into a patch to apply.  Level~2 is
the highest-leverage intervention we measured in calibration, raising
compliance on Claude Haiku~4.5 and Claude
Sonnet~4.6~\cite{anthropic2024claude}.

\paragraph{Level 3: amended input.}
The client changes where the fix appears rather than what it says.  At
Level~2 a cached fix arrives in the memory block as an instruction the
planner is asked to follow, which competes with the task instructions
already there; the planner then has two directives about the same
request and satisfies whichever it read last.  At Level~3 the memory
block presents the fix as an amendment already applied to the input,
so there is one description of the request rather than a request plus
a correction.  This produced a smaller additional gain than Level~2
but eliminated a class of instruction-conflict failures in which the
planner applied some suggestion fields and silently dropped others.
Level~3 subsumes Level~2: every Level~3 configuration carries the
Level~2 sentence as well, which is why the experiments hold the pair
fixed rather than varying them separately
(\S\ref{sec:setup}).

% ------------------------------------------------------------------
\subsection{Two Ladders: Server Levels, Client Arms}
\label{sec:two-ladders}

The protocol levels are the \emph{server's} design decisions: what
information to attach to the response.  They say nothing about what a
client does with it.  A client that receives a version stamp may
ignore it, may flush its whole cache on any change, or may evict one
row.  Those are the \emph{client's} design decisions, and the paper
varies them along a second ladder, the \emph{A-ladder}, whose rungs
are the experimental arms.  Results are reported per arm, so the
A-ladder is the axis of every table in \S\ref{sec:results}.

Six arms of the same client run against the same server.  A0 keeps no
memory and re-derives every constraint; it is the cost baseline.  A1
stores what it learns and injects it later, with no channel by which
the server can report that a stored item has become false.  A2 adds
invalidation at table granularity, so a version bump drops everything
the client holds from that table.  A2D adds invalidation at row
granularity: the server names the row that changed, and the client
drops that row and restamps the rest.  A2DC and A2DG extend A2D with
consumer notification and with propagation along the dependency
graph.  A2DC is a purely client-side strategy: when the client evicts
a row, it notifies downstream consumers that depended on it.  No
additional server output is required, so A2DC activates the same
protocol levels as A2D.  A seventh variant, A2DO, replaces the diff
channel with a stream of rule declarations and is used only on the
policy-rule stream.  Table~\ref{tab:arm-levels} pairs each arm with
the server levels it consumes.

All arms run with the consumption channel fixed at L3
(\S\ref{sec:consumption-levels}), which subsumes L2.  The A-ladder
therefore varies the \emph{invalidation channel} while holding the
\emph{compliance channel} constant.  L2 is not isolated in this
evaluation because L3 builds on it: every L3 arm inherits the ``merge
parameters verbatim'' instruction that defines L2.  The L2-specific
compliance lift was measured in earlier calibration
runs.\footnote{The calibration runs that isolated L2 (a +56\,pp
  compliance lift on two Claude models) predate the experiments
  reported here and are not archived alongside
  Tables~\ref{tab:ladder}--\ref{tab:tokens}.}

The ``Diff'' column in Table~\ref{tab:arm-levels} marks a server
mechanism that does not have its own protocol level: the reload
endpoint returns row-level diffs (added, removed, changed keys)
alongside the version bump.  This is not a new field on the regular
response wire format; it is an operational refinement of the reload
endpoint that makes L1's version stamps actionable at row
granularity.  A2 uses only the version stamp and evicts every entry
from the affected table.  A2D uses the diff to evict only the changed
row and restamp the rest.  The distinction between A2 and A2D is the
client's strategy for the same server output.

\begin{table}[t]
\centering
\caption{Coverage matrix: experimental arms against protocol levels,
  grouped by the savings-equation factor each level engineers
  (Equation~\ref{eq:savings}).  \ding{51} = active in that arm;
  (\ding{51}) = exercised only inside a bundled arm, not isolated.
  All arms shown hold the compliance channel fixed at L2/L3.
  L1 straddles both factors: its structured format aids compliance,
  and its version stamps enable invalidation.  L0 (raw feedback) has
  no arm: every configuration in the evaluation emits at least L1.
  A0 (no memory) consumes no contract field and is the cost baseline
  of Tables~\ref{tab:ladder} and~\ref{tab:tokens}, so it is omitted
  from the matrix.  The A-ladder varies the validity channel; empty
  cells mark untested combinations.}
\label{tab:arm-levels}
\footnotesize
\begin{tblr}{
  colspec = {l cc ccccc l},
  rowsep = 2pt,
  colsep = 3pt,
  row{1} = {font=\bfseries\footnotesize, bg=tblHeaderTint},
  row{2} = {font=\bfseries\footnotesize, bg=tblHeaderTint},
  row{4,6,8} = {bg=tblRowAlt},
  cell{1}{2} = {c=2}{c},
  cell{1}{4} = {c=5}{c},
  hline{1,Z} = {0.8pt},
  hline{2} = {2-3}{0.4pt},
  hline{2} = {4-8}{0.4pt},
  hline{3} = {0.5pt},
}
 & {Compliance} & & {Validity} & & & & & \\
{Arm} & {L2} & {L3} & {L1} & {L4} & {L5} & {L6} & {Diff} & {What the arm adds} \\
A1   & \ding{51} & \ding{51} &     &     &     &     &     & Memory, no invalidation \\
A2   & \ding{51} & \ding{51} & \ding{51} &     &     &     &     & Table-granularity stamps \\
A2D  & \ding{51} & \ding{51} & \ding{51} &     &     &     & \ding{51} & + row-level diffs \\
A2DC & \ding{51} & \ding{51} & \ding{51} &     &     &     & \ding{51} & + consumer notification \\
A2DG & \ding{51} & \ding{51} & \ding{51} & (\ding{51}) & \ding{51} &     & \ding{51} & + subgraph propagation \\
A2DO & \ding{51} & \ding{51} & \ding{51} &     &     & \ding{51} & \ding{51} & + rule-declaration oracle \\
\end{tblr}
\end{table}

A1 against A2 measures what any invalidation is worth, and A2 against
A2D measures what \emph{precise} invalidation is worth.  The second
comparison is the one this paper turns on: reporting A2 alone would
describe a protocol that repairs staleness while billing the client
for memory that never went stale.

% ------------------------------------------------------------------
\subsection{The Savings Equation}
\label{sec:savings-equation}

Realized savings from cross-episode memory decompose into two
independent factors as presented in Equation~\ref{eq:savings}.

\paragraph{Validity} is the contract's responsibility.  A cached
suggestion is valid if every table it depends on still has the version
the client recorded.  Validity is model-independent and
vendor-independent: in our experiments, the \texttt{table\_versions}
check produced identical results across seven models, three serving
paths, and all random seeds.  The funding-type and negative-control policy
columns match cell for cell.

\paragraph{Compliance} is the model's responsibility, modulated by the
feedback-consumption protocol (Levels~2--3).  We observed three
regimes:

\begin{itemize}\tightlist
\item Claude Haiku~4.5: 100\% compliance at A2D on the aggregate
  measure (Table~\ref{tab:ladder}), with per-class rates of
  90\% add-a-field, 75\%/58\% rewrite-a-value, and 50\%/67\%
  conditional-rewrite (Table~\ref{tab:taxonomy}).
\item Claude Sonnet~4.6: 81.5--88.9\% aggregate compliance across
  arms,\footnote{A2 through A2DG in Table~\ref{tab:ladder}.} with
  per-class rates of 80\% add-a-field, 30\%/38\% rewrite-a-value,
  and 50\%/67\% conditional-rewrite.
\item Claude Sonnet~5: 0--11\% aggregate compliance.  Sonnet~5
  applied rewrite-a-value hints at 55--60\% and conditional-rewrite
  hints at 50--67\%, but refused add-a-field hints at
  10\%.\footnote{Per-class rates from Table~\ref{tab:taxonomy} at
    A2D.}  We attribute this to \emph{input-schema conservatism}:
  the model treats the request schema as fixed and will not add
  fields the original request did not contain.
\end{itemize}

The factorization in Equation~\ref{eq:savings} matters because it
separates what a contract designer can control (validity, via protocol
levels~1 and~4--6) from what the model vendor controls (compliance,
via instruction following and schema flexibility).  A contract that
achieves perfect validity still delivers zero savings if the planner
ignores the fix.

% ------------------------------------------------------------------
\subsection{Design Decisions}
\label{sec:design-decisions}

\paragraph{Per-table vs.\ global version stamp.}
A single global version stamp is simpler but forces the client to flush
all cached suggestions whenever any table changes.  Per-table stamps
avoid this: a change to \texttt{required\_promo\_policy} does not
invalidate a cached funding-type fix.  The trade-off is that per-table
stamps still over-evict when only some rows within a table changed.
The reload endpoint's structured diff (added/removed/changed keys)
addresses this by letting the client do row-level invalidation.

\paragraph{What the contract costs on the wire.}
The fields are not free, and the per-table choice above is what they
cost.  Measured over the 3{,}387 archived server responses, 2{,}419 of
which carry contract fields, the contract adds a mean of 123~bytes to
a mean 817-byte response, or 15.1\% of payload; among the responses
that carry them, the mean addition is 173~bytes and the maximum
observed is 251.  The distribution is lopsided.  The
\texttt{table\_versions} envelope accounts for 11.7\% of all bytes
sent, at a mean of 190~bytes, while the two per-suggestion fields
together account for 3.2\% (\texttt{cache\_hint} 24~bytes,
\texttt{table} 28~bytes).  Nearly all the overhead is therefore the
per-table dictionary, and it grows with the number of tables the
server has loaded rather than with the number of suggestions it
emits.  A global stamp would shrink the envelope to one string and
give back most of the 15\%, at the over-eviction cost described
above; per-table stamps buy precise invalidation with bytes that scale
in the size of the reference set.  Two scope limits: these are the
Level~1 fields, which every arm in the ladder carries, and the
figures are gross wire cost.  The net token effect of running the
contract is already inside the savings in
Table~\ref{tab:tokens}, which compares each model's billed usage
against its own A0 baseline.

\paragraph{Two-valued cache hint.}
We considered a richer taxonomy (``cacheable for N hours,'' ``cacheable
within session'') but found the distinction is binary in practice:
either the fix comes from server-side data the model cannot derive from
the request alone, or it does not.  Time-based expiry adds complexity
without value when the server already pushes version stamps on every
response.

\paragraph{Row identity by key, not by value.}
Billing vocabulary is low-entropy.  Multiple reference-table rows may
map to the value \texttt{"annual"} or \texttt{"credit"}.  Matching
cached suggestions to table rows by value causes collisions; matching
by key (e.g., \texttt{funding\_type\_policy:enterprise\_annual}) is
unambiguous.

\paragraph{Explicit graph vs.\ implicit reconstruction.}
A client that tracks Level~4 dependency vectors across episodes can
reconstruct the same edges the Level~5 graph declares explicitly. Each
suggestion's \texttt{tables} list reveals which tables co-participate
in a derivation. Over enough episodes, the client builds a partial
graph that converges toward the server's full graph. The two
representations carry equivalent information for suggestion families
the client has observed. The explicit graph adds the families the
client has not seen. This matters because of failure-gating
(\S\ref{sec:disc-detection}): a cached family stops producing errors
and stops exposing its dependency vector, so the implicit graph has
blind spots exactly where memory works best. The explicit graph fills
them. In the evaluation, we measure both to separate their
contributions.

\paragraph{Pre-reload edge walk.}
The \texttt{affected\_downstream} computation must use the knowledge
graph as it existed \emph{before} the reload.  After the reload,
edges originating from the changed node may have disappeared (because
the new data no longer implies those derivations), but the client's
cache still holds suggestions that traversed those edges.  The server
therefore snapshots the graph, applies the new data, diffs, and then
walks the snapshot to find all downstream nodes the client should
invalidate.

\section{Experimental Setup}
\label{sec:setup}

\paragraph{Domains and drift schedule.}
The primary domain is a payments API with four constraint classes over a 36-episode stream. The
governed class needs a promotional code the task input does not carry, so its hint is an
\emph{add-a-field} hint. The funding class needs a payment token selected by a condition on the
request, a \emph{conditional-rewrite} hint. A negative-control class depends on rows no scheduled
event touches, and a fourth class carries no constraint. Two events run on a fixed schedule: after
episode 11 the code table rotates, which is breaking, so exactly one eviction is correct; after
episode 23 the token table takes a benign version bump, where zero evictions and one restamp are
correct. The second domain is a recipe API with a four-rung dietary ladder over 48 episodes. It
shares the harness, the arms and the scoring, and it supplies the third hint shape, a
\emph{rewrite-a-value} hint that replaces an ingredient the task already names.

\paragraph{Models and serving stacks.}
Seven models on three serving paths. \texttt{claude-haiku-4-5}, \texttt{claude-sonnet-4-6} and
\texttt{claude-sonnet-5} are served by Anthropic directly. \texttt{gpt-5.4-mini} and
\texttt{deepseek-v4-flash} reach their vendors through a third-party API gateway, and
\texttt{gpt-5-mini} and \texttt{gemini-3.5-flash} through a translation layer that lets a harness
speaking one vendor's message protocol call two others. Every run opens with a preflight probe that
records the requested and the served model identifier: across all 238 archived run artifacts the two
agree, with no mismatch and none missing a probe. The model count is not there to rank models. It
separates what the protocol does from what any one client happens to do, and the three paths test
whether the result survives a change of delivery route.

\paragraph{Metrics.}
Compliance is the first-try rate over injection-eligible episodes of the governed class. Eviction
precision is the reciprocal of a run's eviction count. Retries are summed over the stream, and
first-try rate is also reported per constraint class. Token counts come from each vendor's billed
usage field. The three serving paths do not bill on a common basis, so token counts are compared
only within a model against that model's own A0 and never across paths. An absolute rate for a
model is a rate for its delivery path, and every comparison this paper draws is between arms of one
model on one path.

\subsection{Measurement Hygiene}
\label{sec:hygiene}

\paragraph{What compliance excludes, and why.}
The first governed episode after each rotation is a slot in which every memory arm still injects
the pre-rotation code, because no arm can know the code changed until the server says so. A retry
there prices staleness rather than non-compliance, and scoring those slots would penalize A1, A2 and
A2D for the same event and flatten the differences among them. They are excluded by a fixed
episode-index list rather than by inspection of the results. Two choices here move the headline
numbers and we state both. Widening the slot boundary by one episode raises A1 compliance. Treating
a table rotation as warranting a cleared table, rather than the single changed row the server's own
diff names, would score A2 at 1.00 precision instead of 0.25. Both are fixed by the pre-registered
design rather than chosen after the runs.

\paragraph{An exhausted run can imitate a precise one.}
Eviction precision divides by the number of evictions a run performs. A run that exhausts its retry
budget early never reaches the later drift events, performs fewer evictions, and reports a higher
precision than a run that completed. \texttt{gemini-3.5-flash} reports 1.00 at A2 for exactly that
reason, and its restamp count reads 0, 0, 1 against a reference of 3 because the episodes carrying
the diff observation do not complete. Any evaluation that scores an agent on event counts inherits
this hazard, since the broken run looks disciplined. We report event-count metrics next to the
number of completed episodes so the denominator stays visible.

\section{Results}
\label{sec:results}

%% AUTO-GENERATED by scratchpad/gen_tables.py. DO NOT hand-edit numbers.
\begin{table*}[t]
\centering
\caption{The A-ladder on the drifted stream, three seeds per cell, row blocks by serving stack.
Compliance excludes the stale-injection slots (Section~\ref{sec:hygiene}) and is undefined at A0.
Eviction precision is 1.00 at A2D on every model, so only A2 is shown; seed standard deviation never
exceeds 18.1 points. Gains of A2D over A1 are in the \colorbox{DeltaBg}{$\Delta$ column}, shaded by
the unrounded gain; \underline{underlined} is the best A2D cell in each stack.}
\label{tab:ladder}
\footnotesize
\begin{tblr}{
  colspec = {l ccccc c c ccc},
  rowsep = 2pt,
  colsep = 3pt,
  column{1} = {font=\ttfamily},
  row{1} = {font=\bfseries\footnotesize, bg=tblHeaderTint},
  row{2} = {font=\bfseries\footnotesize, bg=tblHeaderTint},
  row{4} = {bg=tblRowAlt},
  row{7} = {bg=tblRowAlt},
  row{9} = {bg=tblRowAlt},
  cell{1}{1} = {r=2}{valign=m},
  cell{1}{2} = {c=5}{c, bg=SearchBg},
  cell{1}{7} = {bg=DeltaBg},
  cell{1}{8} = {bg=ScienceBg},
  cell{1}{9} = {c=3}{c, bg=AgenticBg},
  hline{1,Z} = {0.8pt},
  hline{2} = {2-6}{0.4pt},
  hline{2} = {7-7}{0.4pt},
  hline{2} = {8-8}{0.4pt},
  hline{2} = {9-11}{0.4pt},
  hline{3} = {0.5pt},
  hline{6,8} = {0.3pt},
}
{Model} & {Compliance (\%)} & & & & & {$\Delta$} & {Prec.} & {Post-drift funding (\%)} & & \\
 & {A1} & {A2} & {A2D} & {A2DC} & {A2DG} & {A2D$-$A1} & {A2} & {A1} & {A2} & {A2D} \\
claude-haiku-4-5 & 44.4 & 100.0 & \underline{100.0} & 100.0 & 100.0 & \SetCell{bg=darkgreen!56} \gain{55.6} & 0.25 & 66.7 & 0.0 & 66.7 \\
claude-sonnet-4-6 & 25.9 & 81.5 & 88.9 & 81.5 & 88.9 & \SetCell{bg=darkgreen!63} \gain{63.0} & 0.25 & 66.7 & 0.0 & 66.7 \\
claude-sonnet-5 & 0.0 & 3.7 & 11.1 & 0.0 & 3.7 & \SetCell{bg=darkgreen!11} \gain{11.1} & 0.25 & 66.7 & 0.0 & 66.7 \\
gpt-5.4-mini & 100.0 & 100.0 & \underline{100.0} & 100.0 & 100.0 & 0.0 & 0.25 & 66.7 & 22.2 & 66.7 \\
deepseek-v4-flash & 3.7 & 7.4 & 3.7 & 18.5 & 0.0 & 0.0 & 0.31 & 66.7 & 66.7 & 55.6 \\
gpt-5-mini & 33.3 & 100.0 & \underline{100.0} & {--} & {--} & \SetCell{bg=darkgreen!67} \gain{66.7} & 0.25 & 66.7 & 0.0 & 66.7 \\
gemini-3.5-flash & 29.6 & 63.0 & 77.8 & {--} & {--} & \SetCell{bg=darkgreen!48} \gain{48.2} & 1.00 & 0.0 & 0.0 & 11.1 \\
\end{tblr}
\end{table*}

\begin{figure}[t]
\centering
\includegraphics[width=0.9\columnwidth]{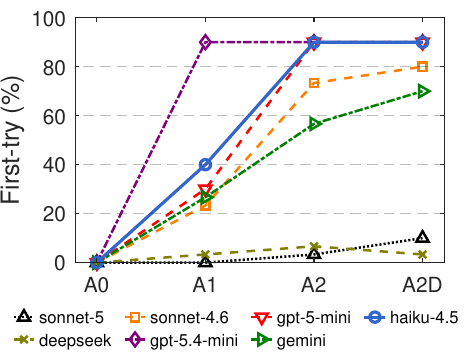}
\caption{The A-ladder on the drifted stream, seven models, three seeds:
first-try rate on the governed class.}
\label{fig:governed}
\end{figure}

\begin{figure}[t]
\centering
\includegraphics[width=0.9\columnwidth]{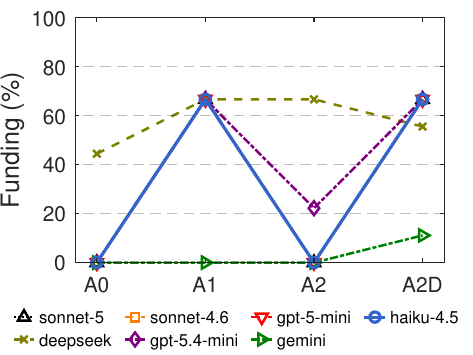}
\caption{Post-drift first-try on the funding class, which no scheduled
event targets. The arm that raises Figure~\ref{fig:governed} is the arm
that empties this one.}
\label{fig:funding}
\end{figure}

The evaluation is 250 arm-level runs over 9,432 episodes, three seeds per drifted cell, seven models,
three serving paths and two domains. Table~\ref{tab:ladder} is the grid; Figures~\ref{fig:governed}
and~\ref{fig:funding} plot the four arms all seven models share.

\paragraph{Row-Level Invalidation Is the Arm That Pays.}
Compliance rises from A1 to A2D on five of seven models, by 66.7 points on \texttt{gpt-5-mini},
63.0 on \texttt{claude-sonnet-4-6}, 55.6 on \texttt{claude-haiku-4-5}, 48.2 on
\texttt{gemini-3.5-flash} and 11.1 on \texttt{claude-sonnet-5}. Retries fall with it: over 36 episodes \texttt{claude-haiku-4-5} and
\texttt{gpt-5-mini} both go from 27 at A0 to 7 at A2D, and \texttt{gemini-3.5-flash} from 82.7 to
55.0. The ladder then stops paying. Consumer notification and graph propagation move compliance between $-11.1$ and
$+14.8$ points with no consistent sign, and since nine episodes are eligible for the score, one
episode is worth 11.1 points and both extremes belong to models within two episodes of a floor or a
ceiling. One model reaches 100.0\% compliance at A1, before any invalidation exists, so the ladder
has nothing left to buy there on the compliance axis; \texttt{gpt-5.4-mini} still moves from 0.25 to
1.00 eviction precision and recovers post-drift funding from 22.2\% to 66.7\%. Compliance and
precision are separate purchases, and the model that saturates the first shows it.

\paragraph{Table-Level Invalidation Bills a Class That Never Drifted.}
A2 clears an entire table on a version bump. That is correct for the rotated code table and it
destroys the funding entry stored beside it. Post-drift funding first-try falls from 66.7\% at A1
to 0.0\% at A2 and returns to 66.7\% at A2D on four of seven models, and a fifth,
\texttt{gpt-5.4-mini}, falls to 22.2\% and returns to the same 66.7\%;
Figure~\ref{fig:funding} is that collapse and recovery. Eviction precision states the same fact without the client in the loop, at 0.25 for
table granularity and exactly 1.00 for row granularity on every model. Two cells carry caveats
rather than readings. \texttt{deepseek-v4-flash} evicts three or four times where the others evict
four, and the seed that skips an eviction keeps its funding entry, which is why its funding column
does not collapse. \texttt{gemini-3.5-flash} reads 1.00 precision at A2 for the starvation reason of
Section~\ref{sec:hygiene}, not because it evicted well. The same starvation keeps its post-drift
funding at 0.0\% from A0 through A2, with 11.1\% at A2D.

%% AUTO-GENERATED by scratchpad/gen_tables.py. DO NOT hand-edit numbers.
\begin{table}[t]
\centering
\caption{First-try rate (\%) by constraint class, three seeds per cell: ten governed episodes,
three funding episodes after the benign bump, six control episodes no scheduled event touches. A
fourth class carries no constraint and reads 100.0 in all fourteen cells, so it is omitted.}
\label{tab:classes}
\footnotesize
\begin{tblr}{
  colspec = {l cc cc cc},
  rowsep = 2pt,
  colsep = 4.5pt,
  column{1} = {font=\ttfamily},
  row{1} = {font=\bfseries\footnotesize, bg=tblHeaderTint},
  row{2} = {font=\bfseries\footnotesize, bg=tblHeaderTint},
  row{4} = {bg=tblRowAlt},
  row{7} = {bg=tblRowAlt},
  row{9} = {bg=tblRowAlt},
  cell{1}{1} = {r=2}{valign=m},
  cell{1}{2} = {c=2}{c, bg=SearchBg},
  cell{1}{4} = {c=2}{c, bg=AgenticBg},
  cell{1}{6} = {c=2}{c, bg=MathBg},
  hline{1,Z} = {0.8pt},
  hline{2} = {2-3}{0.4pt},
  hline{2} = {4-5}{0.4pt},
  hline{2} = {6-7}{0.4pt},
  hline{3} = {0.5pt},
  hline{6,8} = {0.3pt},
}
{Model} & {Add-a-field} & & {Cond.-rewrite} & & {Neg.\ control} & \\
 & {A1} & {A2D} & {A1} & {A2D} & {A1} & {A2D} \\
claude-haiku-4-5 & 40.0 & 90.0 & 66.7 & 66.7 & 83.3 & 83.3 \\
claude-sonnet-4-6 & 23.3 & 80.0 & 66.7 & 66.7 & 83.3 & 83.3 \\
claude-sonnet-5 & 0.0 & 10.0 & 66.7 & 66.7 & 83.3 & 83.3 \\
gpt-5.4-mini & 90.0 & 90.0 & 66.7 & 66.7 & 83.3 & 83.3 \\
deepseek-v4-flash & 3.3 & 3.3 & 66.7 & 55.6 & 83.3 & 83.3 \\
gpt-5-mini & 30.0 & 90.0 & 66.7 & 66.7 & 83.3 & 83.3 \\
gemini-3.5-flash & 26.7 & 70.0 & 0.0 & 11.1 & 44.4 & 55.6 \\
\end{tblr}
\end{table}

\paragraph{The Untargeted Classes Do Not Move, and Governance Costs When Nothing Drifts.}
Table~\ref{tab:classes} carries the controls. All 1,026 no-constraint episodes across the 114 runs
of the drifted stream are first-try, at every arm including A0. The negative-control class holds
at 83.3\% across every memory arm on six of seven models, the exception being the model with a
failure tail, and at A0 that same class drops to 0.0\% on five models, so memory alone is worth 83.3
points there before any invalidation. The bill for that comes due on the undrifted control stream,
where the ladder has nothing to invalidate: \texttt{claude-sonnet-4-6} scores 100.0\% compliance at
A1 and 54.5\% at A2, with A2D at 81.8\% and A2DG at 90.9\%, while \texttt{claude-haiku-4-5} and
\texttt{gpt-5.4-mini} stay at or near 100.0 throughout. The size of the bill is a property of the
client, but it is charged whether or not anything went stale.

\begin{figure}[t]
\centering
\includegraphics[width=0.9\columnwidth]{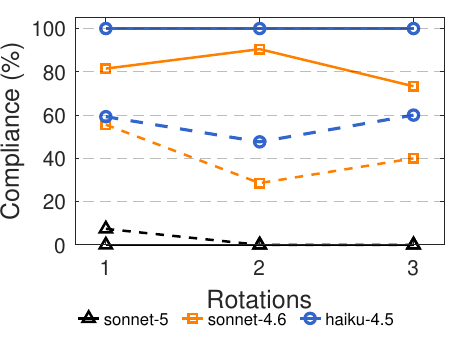}
\caption{Compliance against the number of table rotations packed into
one 36-episode budget, A2D solid against A1 dashed, for the three
models present in both domains.}
\label{fig:frequency}
\end{figure}

\begin{figure}[t]
\centering
\includegraphics[width=0.9\columnwidth]{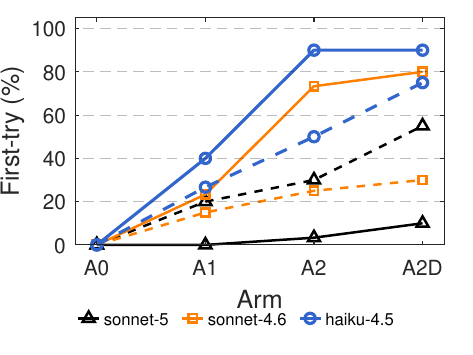}
\caption{First-try rate on the class the drift event targets, payments
solid against recipes dashed, one color per model.}
\label{fig:domain}
\end{figure}

\paragraph{The Gain Holds Across Drift Rate, Domain and Serving Path.}
Figure~\ref{fig:frequency} runs one, two and three rotations through the same episode
budget.
\texttt{claude-haiku-4-5} holds 100.0\% compliance at A2D at every rotation count while its A1
compliance moves between 47.6\% and 60.0\%, and \texttt{claude-sonnet-4-6} holds 73.3\% to 90.5\% at
A2D against 28.6\% to 55.6\% at A1. Retries at A2D stay near 7 and near 9 respectively across all
three rates, so the protocol absorbs the added drift without charging for it. The second domain
reproduces the direction with a smaller margin, and Figure~\ref{fig:domain} puts the two side by side: at A2D
the class the drift event targets rises 48.3 points over A1 on \texttt{claude-haiku-4-5}, 35.0 on
\texttt{claude-sonnet-5} and 15.0 on \texttt{claude-sonnet-4-6}. That figure also separates the client
from the domain, since \texttt{claude-sonnet-5} ends at 10.0\% in payments and 55.0\% in recipes: the
client that opts out of one hint shape climbs the same ladder in the other. A third stream replaces table constraints with a policy rule set, and there the
per-episode retry vector is identical position by position across \texttt{claude-haiku-4-5},
\texttt{claude-sonnet-4-6} and \texttt{claude-sonnet-5} within each arm, at 42 attempts for A2D,
44 for A2DC and 45 for A2DO. The nine artifacts are distinct runs with distinct payloads and token
totals spanning 243k to 390k, so the identity is not a duplicated file: on that stream the arm
fixes the retry ledger and the choice of model does not move it. Table~\ref{tab:pricesheet}
prices the detection policies on that stream (\S\ref{sec:disc-detection}).

\begin{table}[t]
\centering
\caption{Detection policies on the rule stream: 36 episodes, one rule
edge removed after episode~11, one added after episode~23. Every count
is identical on the three models (\texttt{claude-haiku-4-5},
\texttt{claude-sonnet-4-6}, \texttt{claude-sonnet-5}); the token
column is the per-model range against lazy vectors;
\S\ref{sec:disc-detection} carries the exact per-model values.}
\label{tab:pricesheet}
\footnotesize
\begin{tblr}{
  colspec = {l c c c c c c},
  rowsep = 2pt,
  colsep = 2pt,
  row{1} = {font=\bfseries\footnotesize, bg=tblHeaderTint},
  row{3,5} = {bg=tblRowAlt},
  hline{1,Z} = {0.8pt},
  hline{2} = {0.5pt},
}
{Policy} & {First-try} & {Retries} & {Evict} & {Restamp} & {Alert} & {$\Delta$ tok.\ (\%)} \\
Lazy (\texttt{A2DC}) & 28 & 8 & 2 & 3 & 13/25 & {--} \\
Stamp-only (\texttt{A2DR}) & 22 & 14 & 4+5 & 0 & 12/24 & +14.0 to +14.3 \\
Stamp+vec.\ (\texttt{A2DRC}) & 28 & 8 & 2 & 3 & 12/24 & $-$0.11 to +0.20 \\
Oracle (\texttt{A2DO}) & 27 & 9 & 3 & 3 & 12/24 & +2.42 to +2.45 \\
\end{tblr}
\end{table}

%% AUTO-GENERATED by scratchpad/gen_tables.py. DO NOT hand-edit numbers.
\begin{table}[t]
\centering
\caption{Token and retry cost on the drifted stream, three seeds per cell, row blocks by serving
stack. Saving is measured within a model against its own A0, which is 0.0 by construction. The
\colorbox{DeltaBg}{$\Delta$ column} is what row granularity adds over naive memory;
\underline{underlined} is the best A2D saving in each stack. Savings are net of the
contract's own wire cost, which is 15.1\% of response payload
(\S\ref{sec:design-decisions}).}
\label{tab:tokens}
\footnotesize
\begin{tblr}{
  colspec = {l cc c ccc},
  rowsep = 2pt,
  colsep = 4.5pt,
  column{1} = {font=\ttfamily},
  row{1} = {font=\bfseries\footnotesize, bg=tblHeaderTint},
  row{2} = {font=\bfseries\footnotesize, bg=tblHeaderTint},
  row{4} = {bg=tblRowAlt},
  row{7} = {bg=tblRowAlt},
  row{9} = {bg=tblRowAlt},
  cell{1}{1} = {r=2}{valign=m},
  cell{1}{2} = {c=2}{c, bg=SearchBg},
  cell{1}{4} = {bg=DeltaBg},
  cell{1}{5} = {c=3}{c, bg=AgenticBg},
  hline{1,Z} = {0.8pt},
  hline{2} = {2-3}{0.4pt},
  hline{2} = {4-4}{0.4pt},
  hline{2} = {5-7}{0.4pt},
  hline{3} = {0.5pt},
  hline{6,8} = {0.3pt},
}
{Model} & {Token saving (\%)} & & {$\Delta$} & {Retries per stream} & & \\
 & {A1} & {A2D} & {A2D$-$A1} & {A0} & {A1} & {A2D} \\
claude-haiku-4-5 & 22.3 & \underline{32.5} & \SetCell{bg=darkgreen!41} \gain{10.2} & 27.0 & 13.0 & 7.0 \\
claude-sonnet-4-6 & 18.6 & 29.2 & \SetCell{bg=darkgreen!42} \gain{10.6} & 27.0 & 14.7 & 8.3 \\
claude-sonnet-5 & 5.3 & 8.7 & \SetCell{bg=darkgreen!14} \gain{3.4} & 32.3 & 27.7 & 25.3 \\
gpt-5.4-mini & 28.2 & \underline{31.0} & \SetCell{bg=darkgreen!11} \gain{2.9} & 24.0 & 7.0 & 7.0 \\
deepseek-v4-flash & 7.1 & 6.0 & -1.0 & 20.0 & 16.0 & 16.7 \\
gpt-5-mini & 16.4 & \underline{32.1} & \SetCell{bg=darkgreen!63} \gain{15.7} & 27.0 & 14.0 & 7.0 \\
gemini-3.5-flash & 10.3 & 22.5 & \SetCell{bg=darkgreen!49} \gain{12.2} & 82.7 & 68.3 & 55.0 \\
\end{tblr}
\end{table}

\paragraph{What Row Granularity Adds Is Priced by Compliance.}
Table~\ref{tab:tokens} prices the ladder. On the five models that draw on their memory, naive
memory alone recovers 10.3\% to 28.2\% of a model's own A0 token cost, so most of the headline saving is bought by holding a memory at all. Row
granularity adds 10.2 and 10.6 points on the two compliant Anthropic models, 15.7 on
\texttt{gpt-5-mini} and 12.2 on \texttt{gemini-3.5-flash}, and the size of that increment follows the
client rather than the protocol: the model with the largest compliance gain also has the largest
token gain, while \texttt{claude-sonnet-5} adds 3.4 points and \texttt{deepseek-v4-flash} loses 1.0,
the two clients that decline to draw on what they store. \texttt{gpt-5.4-mini} bounds the rule from
the other side, buying 2.9 points more because its compliance and its retry count are both already at
the floor at A1.

%% AUTO-GENERATED by scratchpad/gen_tables.py. DO NOT hand-edit numbers.
\begin{table}[t]
\centering
\caption{First-try rate (\%) at A2D by hint shape, both domains, three seeds per cell. Only the
opt-out model orders the three shapes; on the two models that use memory the ordering inverts, so no
hint shape is intrinsically harder to apply.}
\label{tab:taxonomy}
\footnotesize
\begin{tblr}{
  colspec = {l c cc cc},
  rowsep = 2pt,
  colsep = 4.5pt,
  column{1} = {font=\ttfamily},
  row{1} = {font=\bfseries\footnotesize, bg=tblHeaderTint},
  row{2} = {font=\bfseries\footnotesize, bg=tblHeaderTint},
  row{4} = {bg=tblRowAlt},
  cell{1}{1} = {r=2}{valign=m},
  cell{1}{2} = {bg=SearchBg},
  cell{1}{3} = {c=2}{c, bg=MathBg},
  cell{1}{5} = {c=2}{c, bg=AgenticBg},
  hline{1,Z} = {0.8pt},
  hline{2} = {2-2}{0.4pt},
  hline{2} = {3-4}{0.4pt},
  hline{2} = {5-6}{0.4pt},
  hline{3} = {0.5pt},
}
{Model} & {Add-a-field} & {Rewrite-a-value} & & {Cond.-rewrite} & \\
 & {promo} & {celiac} & {incompat.} & {pre} & {post} \\
claude-haiku-4-5 & 90.0 & 75.0 & 58.3 & 50.0 & 66.7 \\
claude-sonnet-4-6 & 80.0 & 30.0 & 38.3 & 50.0 & 66.7 \\
claude-sonnet-5 & 10.0 & 55.0 & 60.0 & 50.0 & 66.7 \\
\end{tblr}
\end{table}

\paragraph{Compliance Tracks the Client, Not the Constraint.}
Two clients opt out. \texttt{claude-sonnet-5} and \texttt{deepseek-v4-flash} sit near zero
compliance at every arm while completing the stream, paying in re-derivation retries instead of
memory. Since savings are the product of validity and compliance, a client whose compliance is zero
returns nothing however precise the invalidation, which is the 8.7\% against 32.5\% of
Table~\ref{tab:tokens}. The opt-out is selective
rather than general. Table~\ref{tab:taxonomy} sorts the same runs by the shape of the server's hint:
on \texttt{claude-sonnet-5}, add-a-field hints are applied 10.0\% of the time, rewrite-a-value hints
55.0\% and 60.0\%, and conditional-rewrite hints 50.0\% and 66.7\%, the last being the rate the
compliant models reach. On \texttt{claude-haiku-4-5} and \texttt{claude-sonnet-4-6} the ordering
inverts, with add-a-field at 90.0\% and 80.0\% against conditional-rewrite at 66.7\%, so no hint
shape is intrinsically harder to apply. Failures are rare and concentrated the same way. Of 9,432
episodes, 159 do not complete and all 159 stop at the fifth attempt; five of the seven models never
fail. \texttt{claude-sonnet-5} accounts for 23, all on the governed class and skewed to one task
family, including one at A0, which rules out injected content as the trigger.
\texttt{gemini-3.5-flash} accounts for the other 136, of which 104 belong to the funding family and
32 to the negative control, and its failure count falls monotonically along the ladder from 41 at A0
to 34, 32 and 29.

\section{Discussion}
\label{sec:discussion}

\subsection{What the contract can and cannot engineer}
\label{sec:disc-reach}

The invalidation contract engineers validity reliably. Across seven
models, three serving paths, and all random seeds, the governed protocol
maintains perfect eviction precision at
A2D.\footnote{Eviction precision is 1.00 at A2D on all seven
  models (Table~\ref{tab:ladder}). At A2 it reads 0.25--0.31 on six
  models; the seventh reads 1.00 through the starvation artifact
  described in Section~\ref{sec:hygiene}.} Sonnet~5 holds 1.00
eviction precision while its compliance is zero. It keeps its
cache correct and never draws from it. Validity is
vendor-independent because it depends only on the server's version
stamp, not on the model's willingness to act.

Compliance is harder to engineer. Row-level invalidation (A2D over A1)
is the strongest compliance lever in the experiments reported here, raising
compliance by 55.6 to 66.7 percentage points on three
models.\footnote{Table~\ref{tab:ladder}: Claude Haiku~4.5 (+55.6\,pp),
  Claude Sonnet~4.6 (+63.0\,pp), GPT-5-mini (+66.7\,pp).}
But no protocol level can cross what we call the \emph{input-schema
conservatism} boundary. Sonnet~5 applies rewrite-a-value hints at
55--60\% and conditional-rewrite hints at 50--67\%, but refuses
add-a-field hints at 10\%.\footnote{Per-class rates from
  Table~\ref{tab:taxonomy} at A2D.} The cross-class ordering is
consistent across arms and seeds. The contract cannot override this
decision. It is the model choosing which mutation types to trust.

\paragraph{Practical recommendation.}
Before investing in a full memory system for a given model, run a
\emph{compliance preflight}. Inject a handful of episodes with
known-valid fixes covering each action type the deployment needs.
The preflight is cheap (tens of API calls) and reads out the
compliance factor directly. If the model refuses the action types
that matter, no amount of protocol engineering will help.

\subsection{The A0 zero floor}
\label{sec:disc-a0}

All three Anthropic models score 0\% first-try on the governed class
and 0\% on funding at A0 (no memory).\footnote{Direct readings: Figures~\ref{fig:governed}
  and~\ref{fig:funding} include A0, and the A0 column of
  Table~\ref{tab:tokens} shows 27--32 retries per stream on these
  models. Table~\ref{tab:classes} reports A1 and A2D only.}
DeepSeek solves funding tasks at A0 without
memory.\footnote{Figure~\ref{fig:funding}: \texttt{deepseek-v4-flash}
  reads 44.4\% post-drift funding at A0.}  The task is solvable
without memory for some models.

This matters for interpreting the memory lift. Memory does not make an
impossible task possible. It makes a solvable task cheaper by letting
the client skip re-derivation from the server's recovery feedback.
The zero floor is a model property, not a task property. It reflects
the model's ability to extract the answer from the raw API response
without any cached context. Evaluations that do not measure A0 cannot
distinguish ``memory helped'' from ``the task was already easy.''

\subsection{Eager versus lazy detection}
\label{sec:disc-detection}

The detection-policy price sheet (Table~\ref{tab:pricesheet})
carries live readings for all four policies on the rule stream, with
all three models giving the same counts.

Stamp-only detection (checking \texttt{rules\_version} on every
response) alerts at the drift episode but disposes by blanket clear,
4~entries at the first rule event and 5~at the second. The
re-derivation cost is 22~first-try successes against 14~retries, at
14.0--14.3\% more tokens than the lazy arm. The rule-declaration
oracle (A2DO) reads 27~first-try successes and 9~retries, with
3~evictions and 3~restamps, at 2.4\% more tokens.

Stamp-plus-vectors (checking the stamp on every response and
comparing cached dependency vectors on failure) reproduces the lazy
arm's disposition episode for episode, 2~evictions and 3~restamps,
while alerting at the drift episode rather than one after it. The
token difference against the lazy arm is +0.16\%, +0.20\%, and
$-$0.11\% across the three models.

Vector-only detection gives precision 1.00 but is
\emph{failure-gated}. The better the cached memory works, the fewer
failed requests the client observes, and dependency vectors ride
only on \texttt{recovery\_feedback}, which appears only on failure.
A cached family that stops producing errors stops being observable.

Stamp-plus-vectors dominates. The stamp provides zero-lag detection
on every response, success or failure. The vectors provide surgical
disposition, identifying which specific entries are stale. The
combined policy keeps the lazy arm's disposition, alerts one episode
earlier, and costs about 0.2\% in tokens; next to the oracle it reads
one more first-try success at 2.2--2.5\% fewer tokens. The protocol
should carry both.

\subsection{Limitations}
\label{sec:limitations}

\paragraph{Synthetic APIs.}
Both evaluation domains (recipe conversion, Acme billing) are authored
for this study. The reflective layer that emits \texttt{cache\_hint}
and \texttt{recovery\_feedback} is hand-written in each case. We do
not know how the contract transfers to production APIs where the
schema is larger, drift is organic, and the reflective layer must be
retrofitted.

\paragraph{Scale.}
The evaluation covers two domains, seven models, and approximately
9{,}400 episodes across 250 arm-level runs. Reproducibility is
strong (all seeds converge), but the domain count limits claims about
generality.

\paragraph{Single-table dependencies.}
The current protocol (Levels~1--3) tags each suggestion with one
source table. Multi-table dependencies (Level~4, dependency vectors)
are implemented in the server. The A2DG arm
(Table~\ref{tab:ladder}) evaluates graph propagation with LLMs in the
loop, but the compliance effect of dependency vectors alone (Level~4
without the graph) has not been isolated.\footnote{Whether A2DG
  constitutes a full evaluation of Level~4 is a judgment call; the arm
  bundles graph propagation with row-level diffs, so the marginal
  compliance contribution of vectors is not separated. The
  stamp-plus-vectors policy of \S\ref{sec:disc-detection} reads
  dependency vectors without the graph, but at the detection layer
  and on the rule stream only.}

\paragraph{Prompt and task coverage.}
Compliance is measured with one base prompt per domain and one task
family per content class. Prompt sensitivity is plausible but
untested.

\paragraph{Frequency saturation.}
The naive-governed gap stopped widening at three drift rotations. The
pre-registered falsification criterion fired. We predicted the gap
would grow monotonically, and it did not. Episode-level decomposition
is needed to determine whether the plateau reflects a ceiling effect
or an artifact of the rotation schedule.

\subsection{Future work}
\label{sec:future-work}

Five directions follow from the current results.

\paragraph{Dependency vectors and knowledge subgraphs (L4--L6).}
The server can emit derivation edges that link each cached suggestion
to the schema columns it depends on. The client caches the subgraph
and walks edges on invalidation rather than evicting all entries from
the affected table. Step zero, stable-key identity, is implemented.
The pre-reload edge-walk bug we found during development shows why the
server must own the causal structure. The client cannot reconstruct
derivation provenance from the payload alone.

\paragraph{Live API deployment.}
In a production setting, the drift driver becomes the production drift
engine (schema migrations, configuration changes). The grid metrics
become instrumentation (dashboards over eviction precision and
compliance). The trace-based scoring method fingerprints clients
that the API operator does not control.

\paragraph{Drift-policy discovery from client-side analytics.}
The invalidation contract provides reactive validity: the client
evicts stale entries after the server signals a change. A client that
accumulates drift history across many episodes could go further and
learn the server's drift \emph{policy}. Patterns such as periodic
rotations, co-drifting tables, or escalating severity would let the
client anticipate invalidity before the version stamp moves. For
example, a client that observes three quarterly CSM code rotations
could preemptively lower its confidence in cached CSM fixes as the
next quarter approaches. This adds a temporal dimension to the
validity term in the savings equation: instead of reacting to drift,
the client predicts it.

\paragraph{The contract as an attack surface.}
Every mechanism in this paper asks the client to trust the server's
account of its own freshness, and we evaluated that account only under
an honest server.  Three exposures follow, and the first is the one
our own results create.  The compliance channel is engineered to
remove the planner's discretion: Level~2 instructs the model to merge
suggestion parameters verbatim without paraphrasing or omitting
fields, and Level~3 presents them as amendments already applied to the
input.  A suggestion is therefore a parameter-injection channel into
the next request, and the two levels that make the contract work are
the two that remove the planner's opportunity to notice.  Compliance
and credulity are the same measurement, and A2D's 100\% on
\texttt{claude-haiku-4-5} reads either way.  Second, validity is
asserted, not verified.  A server that never bumps a version pins a
stale fix in client memory indefinitely, and the client has no
independent evidence a table moved; the same lever run the other way,
spurious bumps on every response, forces continuous re-derivation and
converts the 29--33\% token saving into an equivalent surcharge, which
is a cheap denial-of-savings attack against a client that cannot
refuse the signal.  Third, the Level~5 graph endpoint publishes
internal derivation structure, and the \texttt{table} field names
server-side tables on every suggestion, so precise invalidation is
bought with schema disclosure.  Signed version stamps, rate-limited
bumps, and a client-side sanity bound on eviction rate would each
address part of this.  None is implemented here, and a threat model
for cross-episode memory is separate work.

\paragraph{Cross-domain compliance isolation.}
A recipe-domain run with an Acme-style patch section in the system
prompt would confirm whether the feedback-consumption protocol
(Level~2) is the compliance lever independent of the domain's payload
shape. If compliance tracks Level~2 regardless of domain, the
protocol is portable. If it tracks domain, the contract needs
domain-specific calibration.

\section{Conclusion}
\label{sec:conclusion}

The invalidation contract is a protocol-layer artifact that keeps
cross-episode memory correct under server-side data drift. Two fields
per suggestion (a cacheability hint and a version stamp) plus a
structured diff on schema reload give the client enough information to
evict stale entries without trial and error.

Validity is the cheap half. Across seven models, three serving paths,
two domains, and approximately 9,400 episodes, every version-stamp
check produced identical results, deterministic by construction and
confirmed at scale. Eviction precision reaches 1.00 at row granularity
on every model under the row-level oracle of \S\ref{sec:hygiene},
while table-level invalidation over-evicts at 0.25 precision and
destroys co-located entries that never drifted. The contract costs
15\% of response payload, almost all of it the per-table version
dictionary.

Compliance is the half that binds. Row-level invalidation raises it by
0 to 66.7 percentage points across the seven models, and by 55.6 to
66.7 on three of them, while two models decline to draw on cached
suggestions at all. Identical wire bytes, opposite outcomes. Model
recency does not predict which: the newest model tested is the least
compliant, refusing fixes that add a field its input schema did not
already contain. No protocol level crosses that boundary, and we do
not expect one to.

The practical consequence is that an API designer can ship the
contract without replicating the benchmark, but cannot assume it will
pay. A compliance preflight, tens of API calls carrying known-valid
fixes, reads out the model factor directly, and it should be run per
model and per action type before the memory is trusted to save
anything.

\section*{Acknowledgements}

This work used AI coding assistants (Claude Code, Anthropic) during
implementation and manuscript preparation. Claude Code assisted with
refactoring policy modules, authoring drift snapshots, implementing
the dependency-vector and knowledge-graph extensions (Levels~4--6),
and writing the translation proxy used for non-Claude model
evaluation. All experimental design decisions, protocol-level
definitions, the savings equation, hypothesis registration, and
result interpretation were made by the human authors.

\bibliographystyle{ACM-Reference-Format}
\bibliography{refs}


\begin{thebibliography}{37}

%%% the Web address.

\ifx \showCODEN    \undefined \def \showCODEN     #1{\unskip}     \fi
\ifx \showISBNx    \undefined \def \showISBNx     #1{\unskip}     \fi
\ifx \showISBNxiii \undefined \def \showISBNxiii  #1{\unskip}     \fi
\ifx \showISSN     \undefined \def \showISSN      #1{\unskip}     \fi
\ifx \showLCCN     \undefined \def \showLCCN      #1{\unskip}     \fi
\ifx \shownote     \undefined \def \shownote      #1{#1}          \fi
\ifx \showarticletitle \undefined \def \showarticletitle #1{#1}   \fi
\ifx \showURL      \undefined \def \showURL       {\relax}        \fi
% invisible to TeX
\providecommand\bibfield[2]{#2}
\providecommand\bibinfo[2]{#2}
\providecommand\natexlab[1]{#1}
\providecommand\showeprint[2][]{arXiv:#2}

\bibitem[{Anthropic}(2024)]%
        {anthropic2024claude}
\bibfield{author}{\bibinfo{person}{{Anthropic}}.}
  \bibinfo{year}{2024}\natexlab{}.
\newblock \bibinfo{title}{The Claude Model Family}.
\newblock
\urldef\tempurl%
\url{https://www.anthropic.com/claude}
\showURL{%
\tempurl}


\bibitem[{Anthropic}(2025)]%
        {mcp2025}
\bibfield{author}{\bibinfo{person}{{Anthropic}}.}
  \bibinfo{year}{2025}\natexlab{}.
\newblock \bibinfo{title}{Model Context Protocol Specification: Tools}.
\newblock


\bibitem[Canedo and Grama(2026)]%
        {canedo2026selfreflective}
\bibfield{author}{\bibinfo{person}{Arquimedes Canedo} {and}
  \bibinfo{person}{Chethan Grama}.} \bibinfo{year}{2026}\natexlab{}.
\newblock \showarticletitle{Self-Reflective APIs: Structure Beats Verbosity for
  AI Agent Recovery}.
\newblock  (\bibinfo{year}{2026}).
\newblock
\showeprint[arxiv]{2606.05037}


\bibitem[Chen et~al\mbox{.}(2024)]%
        {chen2023selfdebugging}
\bibfield{author}{\bibinfo{person}{Xinyun Chen}, \bibinfo{person}{Maxwell Lin},
  \bibinfo{person}{Nathanael Scharli}, {and} \bibinfo{person}{Denny Zhou}.}
  \bibinfo{year}{2024}\natexlab{}.
\newblock \showarticletitle{Teaching Large Language Models to Self-Debug}. In
  \bibinfo{booktitle}{\emph{ICLR}}.
\newblock


\bibitem[Chen et~al\mbox{.}(2023)]%
        {chen2023teval}
\bibfield{author}{\bibinfo{person}{Zehui Chen}, \bibinfo{person}{Weihua Du},
  \bibinfo{person}{Wenwei Zhang}, \bibinfo{person}{Kuikun Liu},
  \bibinfo{person}{Jiangning Liu}, \bibinfo{person}{Miao Zheng},
  {et~al\mbox{.}}} \bibinfo{year}{2023}\natexlab{}.
\newblock \showarticletitle{{T-Eval}: Evaluating the Tool Utilization
  Capability Step by Step}.
\newblock  (\bibinfo{year}{2023}).
\newblock


\bibitem[Dai et~al\mbox{.}(2026)]%
        {dai2026feedbackeval}
\bibfield{author}{\bibinfo{person}{Dekun Dai}, \bibinfo{person}{Mingwei Liu},
  \bibinfo{person}{Anji Li}, \bibinfo{person}{Jialun Cao},
  \bibinfo{person}{Yanlin Wang}, \bibinfo{person}{Chong Wang},
  \bibinfo{person}{Xin Peng}, {and} \bibinfo{person}{Zibin Zheng}.}
  \bibinfo{year}{2026}\natexlab{}.
\newblock \showarticletitle{{FeedbackEval}: A Benchmark for Evaluating Large
  Language Models in Feedback-Driven Code Repair Tasks}.
\newblock  (\bibinfo{year}{2026}).
\newblock


\bibitem[Fensel et~al\mbox{.}(2005)]%
        {wsmo2005}
\bibfield{author}{\bibinfo{person}{Dieter Fensel}, \bibinfo{person}{Holger
  Lausen}, \bibinfo{person}{Axel Polleres}, {et~al\mbox{.}}}
  \bibinfo{year}{2005}\natexlab{}.
\newblock \bibinfo{title}{Web Service Modeling Ontology ({WSMO})}.
\newblock \bibinfo{howpublished}{W3C Member Submission}.
\newblock


\bibitem[Fielding(2008)]%
        {fielding2008hateoas}
\bibfield{author}{\bibinfo{person}{Roy~T. Fielding}.}
  \bibinfo{year}{2008}\natexlab{}.
\newblock \bibinfo{title}{{REST} {APIs} Must Be Hypertext-Driven}.
\newblock


\bibitem[Fielding et~al\mbox{.}(2014)]%
        {rfc7234}
\bibfield{author}{\bibinfo{person}{Roy~T. Fielding}, \bibinfo{person}{Mark
  Nottingham}, {and} \bibinfo{person}{Julian Reschke}.}
  \bibinfo{year}{2014}\natexlab{}.
\newblock \bibinfo{title}{Hypertext Transfer Protocol ({HTTP}/1.1): Caching}.
\newblock \bibinfo{howpublished}{RFC 7234}.
\newblock


\bibitem[Fowler(2010)]%
        {fowler2010maturity}
\bibfield{author}{\bibinfo{person}{Martin Fowler}.}
  \bibinfo{year}{2010}\natexlab{}.
\newblock \bibinfo{title}{Richardson Maturity Model}.
\newblock


\bibitem[Gou et~al\mbox{.}(2024)]%
        {gou2024critic}
\bibfield{author}{\bibinfo{person}{Zhibin Gou}, \bibinfo{person}{Zhihong Shao},
  \bibinfo{person}{Yeyun Gong}, \bibinfo{person}{Yelong Shen},
  \bibinfo{person}{Yujiu Yang}, \bibinfo{person}{Nan Duan}, {and}
  \bibinfo{person}{Weizhu Chen}.} \bibinfo{year}{2024}\natexlab{}.
\newblock \showarticletitle{{CRITIC}: Large Language Models Can Self-Correct
  with Tool-Interactive Critiquing}. In \bibinfo{booktitle}{\emph{ICLR}}.
\newblock


\bibitem[{GraphQL Foundation}(2021)]%
        {graphql2021}
\bibfield{author}{\bibinfo{person}{{GraphQL Foundation}}.}
  \bibinfo{year}{2021}\natexlab{}.
\newblock \bibinfo{title}{{GraphQL} Specification: Introspection and Error
  Handling}.
\newblock


\bibitem[Huang et~al\mbox{.}(2024)]%
        {huang2023cannotcorrect}
\bibfield{author}{\bibinfo{person}{Jie Huang}, \bibinfo{person}{Xinyun Chen},
  \bibinfo{person}{Swaroop Mishra}, \bibinfo{person}{Huaixiu~Steven Zheng},
  \bibinfo{person}{Adams~Wei Yu}, \bibinfo{person}{Xinying Song}, {and}
  \bibinfo{person}{Denny Zhou}.} \bibinfo{year}{2024}\natexlab{}.
\newblock \showarticletitle{Large Language Models Cannot Self-Correct Reasoning
  Yet}. In \bibinfo{booktitle}{\emph{ICLR}}.
\newblock


\bibitem[Kamoi et~al\mbox{.}(2024)]%
        {kamoi2024selfcorrection}
\bibfield{author}{\bibinfo{person}{Ryo Kamoi}, \bibinfo{person}{Sarkar
  Snigdha~Sarathi Das}, \bibinfo{person}{Nianqi Goyal},
  \bibinfo{person}{Pratyay Gupta}, \bibinfo{person}{Yusen Li},
  \bibinfo{person}{Frank~F. Xu}, {and} \bibinfo{person}{Dan Roth}.}
  \bibinfo{year}{2024}\natexlab{}.
\newblock \showarticletitle{When Can {LLMs} Actually Correct Their Own
  Mistakes? {A} Critical Survey of Self-Correction of {LLMs}}.
\newblock \bibinfo{journal}{\emph{Transactions of the ACL}}
  (\bibinfo{year}{2024}).
\newblock


\bibitem[Le et~al\mbox{.}(2026)]%
        {le2026pairbench}
\bibfield{author}{\bibinfo{person}{Cuong~Chi Le}, \bibinfo{person}{Aashish
  Yadavally}, \bibinfo{person}{Son Le-Anh}, {and} \bibinfo{person}{Tien~N.
  Nguyen}.} \bibinfo{year}{2026}\natexlab{}.
\newblock \showarticletitle{Benchmarking Code Improvement with Progressive,
  Adaptive, and Interactive Feedback}.
\newblock  (\bibinfo{year}{2026}).
\newblock


\bibitem[Li et~al\mbox{.}(2023)]%
        {li2023apibank}
\bibfield{author}{\bibinfo{person}{Minghao Li}, \bibinfo{person}{Feifan Song},
  \bibinfo{person}{Bowen Yu}, \bibinfo{person}{Haiyang Yu},
  \bibinfo{person}{Zhoujun Li}, \bibinfo{person}{Fei Huang}, {and}
  \bibinfo{person}{Yongbin Li}.} \bibinfo{year}{2023}\natexlab{}.
\newblock \showarticletitle{{API-Bank}: A Comprehensive Benchmark for
  Tool-Augmented {LLMs}}.
\newblock  (\bibinfo{year}{2023}).
\newblock


\bibitem[Liang et~al\mbox{.}(2023)]%
        {liang2023taskmatrix}
\bibfield{author}{\bibinfo{person}{Yaobo Liang}, \bibinfo{person}{Chenfei Wu},
  \bibinfo{person}{Ting Song}, \bibinfo{person}{Wenshan Wu},
  \bibinfo{person}{Yan Xia}, \bibinfo{person}{Yu Liu}, {et~al\mbox{.}}}
  \bibinfo{year}{2023}\natexlab{}.
\newblock \showarticletitle{{TaskMatrix.AI}: Completing Tasks by Connecting
  Foundation Models with Millions of {APIs}}.
\newblock  (\bibinfo{year}{2023}).
\newblock


\bibitem[Liu et~al\mbox{.}(2024)]%
        {liu2023agentbench}
\bibfield{author}{\bibinfo{person}{Xiao Liu}, \bibinfo{person}{Hao Yu},
  \bibinfo{person}{Hanchen Zhang}, \bibinfo{person}{Yifan Xu},
  \bibinfo{person}{Xuanyu Lei}, \bibinfo{person}{Hanyu Lai},
  \bibinfo{person}{Yu Gu}, {et~al\mbox{.}}} \bibinfo{year}{2024}\natexlab{}.
\newblock \showarticletitle{{AgentBench}: Evaluating {LLMs} as Agents}. In
  \bibinfo{booktitle}{\emph{ICLR}}.
\newblock


\bibitem[Madaan et~al\mbox{.}(2023)]%
        {madaan2023selfrefine}
\bibfield{author}{\bibinfo{person}{Aman Madaan}, \bibinfo{person}{Niket
  Tandon}, \bibinfo{person}{Prakhar Gupta}, \bibinfo{person}{Skyler Hallinan},
  \bibinfo{person}{Luyu Gao}, \bibinfo{person}{Sarah Wiegreffe},
  \bibinfo{person}{Uri Alon}, \bibinfo{person}{Nouha Dziri},
  \bibinfo{person}{Shrimai Prabhumoye}, \bibinfo{person}{Yiming Yang},
  {et~al\mbox{.}}} \bibinfo{year}{2023}\natexlab{}.
\newblock \showarticletitle{Self-Refine: Iterative Refinement with
  Self-Feedback}. In \bibinfo{booktitle}{\emph{NeurIPS}}.
\newblock


\bibitem[Martin et~al\mbox{.}(2004)]%
        {owls2004}
\bibfield{author}{\bibinfo{person}{David Martin}, \bibinfo{person}{Mark
  Burstein}, \bibinfo{person}{Jerry Hobbs}, {et~al\mbox{.}}}
  \bibinfo{year}{2004}\natexlab{}.
\newblock \bibinfo{title}{{OWL-S}: Semantic Markup for Web Services}.
\newblock \bibinfo{howpublished}{W3C Member Submission}.
\newblock


\bibitem[Mialon et~al\mbox{.}(2023)]%
        {mialon2023augmented}
\bibfield{author}{\bibinfo{person}{Gregoire Mialon}, \bibinfo{person}{Roberto
  Dessi}, \bibinfo{person}{Maria Lomeli}, \bibinfo{person}{Christoforos
  Nalmpantis}, \bibinfo{person}{Ram Pasunuru}, \bibinfo{person}{Roberta
  Raileanu}, \bibinfo{person}{Baptiste Roziere}, \bibinfo{person}{Timo Schick},
  {et~al\mbox{.}}} \bibinfo{year}{2023}\natexlab{}.
\newblock \showarticletitle{Augmented Language Models: A Survey}.
\newblock  (\bibinfo{year}{2023}).
\newblock


\bibitem[Nottingham and Wilde(2016)]%
        {rfc7807}
\bibfield{author}{\bibinfo{person}{Mark Nottingham} {and} \bibinfo{person}{Erik
  Wilde}.} \bibinfo{year}{2016}\natexlab{}.
\newblock \bibinfo{title}{Problem Details for {HTTP} {APIs}}.
\newblock \bibinfo{howpublished}{RFC 7807}.
\newblock


\bibitem[Nottingham and Wilde(2023)]%
        {rfc9457}
\bibfield{author}{\bibinfo{person}{Mark Nottingham} {and} \bibinfo{person}{Erik
  Wilde}.} \bibinfo{year}{2023}\natexlab{}.
\newblock \bibinfo{title}{Problem Details for {HTTP} {APIs}}.
\newblock \bibinfo{howpublished}{RFC 9457}.
\newblock


\bibitem[Olausson et~al\mbox{.}(2024)]%
        {olausson2024selfrepair}
\bibfield{author}{\bibinfo{person}{Theo~X. Olausson},
  \bibinfo{person}{Jeevana~Priya Inala}, \bibinfo{person}{Chenglong Wang},
  \bibinfo{person}{Jianfeng Gao}, {and} \bibinfo{person}{Armando
  Solar-Lezama}.} \bibinfo{year}{2024}\natexlab{}.
\newblock \showarticletitle{Is Self-Repair a Silver Bullet for Code
  Generation?}. In \bibinfo{booktitle}{\emph{ICLR}}.
\newblock


\bibitem[{OpenAI}(2025)]%
        {openai2025functioncalling}
\bibfield{author}{\bibinfo{person}{{OpenAI}}.} \bibinfo{year}{2025}\natexlab{}.
\newblock \bibinfo{title}{Function Calling}.
\newblock


\bibitem[{OpenAPI Initiative}(2021)]%
        {openapi2021}
\bibfield{author}{\bibinfo{person}{{OpenAPI Initiative}}.}
  \bibinfo{year}{2021}\natexlab{}.
\newblock \bibinfo{title}{{OpenAPI} Specification 3.x}.
\newblock


\bibitem[Paul et~al\mbox{.}(2023)]%
        {paul2023refiner}
\bibfield{author}{\bibinfo{person}{Debjit Paul}, \bibinfo{person}{Mete
  Ismayilzada}, \bibinfo{person}{Maxime Peyrard}, \bibinfo{person}{Beatriz
  Borber}, \bibinfo{person}{Antoine Bosselut}, {and} \bibinfo{person}{Robert
  West}.} \bibinfo{year}{2023}\natexlab{}.
\newblock \showarticletitle{{REFINER}: Reasoning Feedback on Intermediate
  Representations}.
\newblock  (\bibinfo{year}{2023}).
\newblock


\bibitem[Qin et~al\mbox{.}(2023)]%
        {qin2023toolllm}
\bibfield{author}{\bibinfo{person}{Yujia Qin}, \bibinfo{person}{Shihao Liang},
  \bibinfo{person}{Yining Ye}, \bibinfo{person}{Kunlun Zhu},
  \bibinfo{person}{Lan Yan}, \bibinfo{person}{Yaxi Lu}, \bibinfo{person}{Yankai
  Lin}, {et~al\mbox{.}}} \bibinfo{year}{2023}\natexlab{}.
\newblock \showarticletitle{{ToolLLM}: Facilitating Large Language Models to
  Master 16000+ Real-World {APIs}}.
\newblock  (\bibinfo{year}{2023}).
\newblock


\bibitem[Shinn et~al\mbox{.}(2023)]%
        {shinn2023reflexion}
\bibfield{author}{\bibinfo{person}{Noah Shinn}, \bibinfo{person}{Federico
  Cassano}, \bibinfo{person}{Ashwarya Gopinath}, \bibinfo{person}{Karthik
  Shukla}, \bibinfo{person}{Karthik Narasimhan}, {and} \bibinfo{person}{Shunyu
  Yao}.} \bibinfo{year}{2023}\natexlab{}.
\newblock \showarticletitle{Reflexion: Language Agents with Verbal
  Reinforcement Learning}. In \bibinfo{booktitle}{\emph{NeurIPS}}.
\newblock


\bibitem[Sriram et~al\mbox{.}(2026)]%
        {sriram2026multitool}
\bibfield{author}{\bibinfo{person}{Shyam Sriram}, \bibinfo{person}{Rahul
  Pandita}, \bibinfo{person}{Ganesh Lakshmanan}, \bibinfo{person}{Ashish
  Shamraj}, {and} \bibinfo{person}{Ripon~K. Saha}.}
  \bibinfo{year}{2026}\natexlab{}.
\newblock \showarticletitle{Improving {LLM}-Assisted Secure Code Generation
  through Retrieval-Augmented-Generation and Multi-Tool Feedback}.
\newblock  (\bibinfo{year}{2026}).
\newblock


\bibitem[Stechly et~al\mbox{.}(2023)]%
        {stechly2023gpt4wrong}
\bibfield{author}{\bibinfo{person}{Karthik Stechly}, \bibinfo{person}{Matthew
  Marquez}, {and} \bibinfo{person}{Subbarao Kambhampati}.}
  \bibinfo{year}{2023}\natexlab{}.
\newblock \showarticletitle{{GPT-4} Doesn't Know It's Wrong: An Analysis of
  Iterative Prompting for Reasoning Problems}. In
  \bibinfo{booktitle}{\emph{NeurIPS 2023 FMDM Workshop}}.
\newblock


\bibitem[Valmeekam et~al\mbox{.}(2023)]%
        {valmeekam2023selfcritique}
\bibfield{author}{\bibinfo{person}{Karthik Valmeekam}, \bibinfo{person}{Matthew
  Marquez}, {and} \bibinfo{person}{Subbarao Kambhampati}.}
  \bibinfo{year}{2023}\natexlab{}.
\newblock \showarticletitle{Can Large Language Models Really Improve by
  Self-Critiquing Their Own Plans?}
\newblock  (\bibinfo{year}{2023}).
\newblock


\bibitem[Wang et~al\mbox{.}(2023)]%
        {wang2023voyager}
\bibfield{author}{\bibinfo{person}{Guanzhi Wang}, \bibinfo{person}{Yuqi Xie},
  \bibinfo{person}{Yunfan Jiang}, \bibinfo{person}{Ajay Mandlekar},
  \bibinfo{person}{Chaowei Xiao}, \bibinfo{person}{Yuke Zhu},
  \bibinfo{person}{Linxi Fan}, {and} \bibinfo{person}{Anima Anandkumar}.}
  \bibinfo{year}{2023}\natexlab{}.
\newblock \showarticletitle{Voyager: An Open-Ended Embodied Agent with Large
  Language Models}.
\newblock  (\bibinfo{year}{2023}).
\newblock


\bibitem[Yang et~al\mbox{.}(2024)]%
        {yang2024sweagent}
\bibfield{author}{\bibinfo{person}{John Yang}, \bibinfo{person}{Carlos~E.
  Jimenez}, \bibinfo{person}{Alexander Wettig}, \bibinfo{person}{Kilian
  Lieret}, \bibinfo{person}{Shunyu Yao}, \bibinfo{person}{Karthik Narasimhan},
  {and} \bibinfo{person}{Ofir Press}.} \bibinfo{year}{2024}\natexlab{}.
\newblock \showarticletitle{{SWE}-agent: Agent-Computer Interfaces Enable
  Automated Software Engineering}.
\newblock  (\bibinfo{year}{2024}).
\newblock


\bibitem[Yang et~al\mbox{.}(2023)]%
        {yang2023intercode}
\bibfield{author}{\bibinfo{person}{John Yang}, \bibinfo{person}{Akshara
  Prabhakar}, \bibinfo{person}{Karthik Narasimhan}, {and}
  \bibinfo{person}{Shunyu Yao}.} \bibinfo{year}{2023}\natexlab{}.
\newblock \showarticletitle{{InterCode}: Standardizing and Benchmarking
  Interactive Coding with Execution Feedback}. In
  \bibinfo{booktitle}{\emph{NeurIPS}}.
\newblock


\bibitem[Yang et~al\mbox{.}(2025)]%
        {yang2025agentprotocols}
\bibfield{author}{\bibinfo{person}{Yingxuan Yang} {et~al\mbox{.}}}
  \bibinfo{year}{2025}\natexlab{}.
\newblock \showarticletitle{A Survey of {AI} Agent Protocols}.
\newblock  (\bibinfo{year}{2025}).
\newblock


\bibitem[Zhou et~al\mbox{.}(2024)]%
        {zhou2024webarena}
\bibfield{author}{\bibinfo{person}{Shuyan Zhou}, \bibinfo{person}{Frank~F. Xu},
  \bibinfo{person}{Hao Zhu}, \bibinfo{person}{Xuhui Zhou},
  \bibinfo{person}{Robert Lo}, \bibinfo{person}{Abishek Sridhar},
  \bibinfo{person}{Xianyi Cheng}, {et~al\mbox{.}}}
  \bibinfo{year}{2024}\natexlab{}.
\newblock \showarticletitle{{WebArena}: A Realistic Web Environment for
  Building Autonomous Agents}. In \bibinfo{booktitle}{\emph{ICLR}}.
\newblock


\end{thebibliography}

\end{document}